\documentclass[10pt,conference]{IEEEtran}
\usepackage[font=footnotesize]{subcaption}
\usepackage{graphicx} % for \scalebox
\usepackage{float}
\usepackage[noadjust]{cite}
\usepackage{amsmath,amssymb}
\usepackage[outdir=./]{epstopdf}
\usepackage{afterpage}
\usepackage{optidef}
\usepackage{orcidlink}
\usepackage{soul}
\usepackage{tikz}
\usetikzlibrary{arrows.meta}
\usetikzlibrary{matrix}
\usepackage{xcolor}
\usepackage{pgfplots}
\usetikzlibrary{shapes.geometric, fit, arrows, positioning, backgrounds}
\usepackage{multirow}

\usepackage{dsfont}

\usepackage{comment}
\usepackage{optidef}

\usepackage{colortbl}  % For cell coloring
\usepackage{tcolorbox} % Hatch patterns

\usepackage{bm} % For bold greek letters
\usepackage{siunitx}

\usepackage{algorithm, algorithmic} % For algorithm box

\DeclareMathOperator*\argmax{arg\,max}

\usetikzlibrary{patterns}
\usepgfplotslibrary{groupplots}

\usepackage{hyperref}
 \hypersetup{
     pdfpagemode=UseNone,
     pdfpagelabels=false,
     bookmarks=false,
    colorlinks=false,
 	linkbordercolor=white,
 	urlbordercolor=white,
 	pdfborder={0 0 0}
 }  
\newcommand*{\marksize}{2.0} %2.6

\newlength\figH
\newlength\figW
\newlength\figWsmall
\usepackage[firstpage=true]{background}
\SetBgContents{\textcolor{black}{\footnotesize This work has been submitted to the IEEE for possible publication. Copyright may be transferred without notice, after which this version may no longer be accessible.}}% Set contents
\SetBgPosition{current page.south}% Select location
\SetBgVshift{0.80cm}% Add vertical shift (results in a shift in x direction due to rotation)
\SetBgOpacity{1.0}% Select opacity
\SetBgAngle{0.0}% Select rotation of logo
\SetBgScale{1.0}% Select scale factor of logo

\makeatletter
\renewcommand{\Hy@writebookmark}[6][]{}
\makeatother
\begin{document}
%\IEEEoverridecommandlockouts
%\bstctlcite{IEEEexample:BSTcontrol}

\IEEEoverridecommandlockouts % Add this to enable \thanks command

\title{Semantic-Aware Task Clustering for Constructive and Cooperative Multi-Tasking \thanks{
This work was partially supported by the Deutsche Forschungsgemeinschaft (DFG, German Research Foundation) under Germany's Excellence Strategy – EXC-3036 The Martian Mindset, project number: 533607631, and by the DFG under project number: 518671822.} \vspace{-1em}
} %and by the Deutsche Forschungsgemeinschaft (DFG, German Research Foundation) -- 518671822.
\author{
\IEEEauthorblockN{Ahmad Halimi Razlighi\,\orcidlink{0009-0006-3826-832X}, Maximilian H. V. Tillmann\,\orcidlink{0009-0000-6548-4278}, Edgar Beck\,\orcidlink{0000-0003-2213-9727}, Bho Matthiesen\IEEEauthorrefmark{2}\,\orcidlink{0000-0002-4582-3938}, and Armin Dekorsy\,\orcidlink{0000-0002-5790-1470}}

\IEEEauthorblockA{Department of Communications Engineering, University of Bremen, Germany}
\IEEEauthorblockA{\IEEEauthorrefmark{2}Communication Systems Group, Paderborn University, 33098 Paderborn, Germany}

\IEEEauthorblockA{E-mails:\{halimi, tillmann, beck, dekorsy\}@ant.uni-bremen.de, \IEEEauthorrefmark{2}E-mail: matthiesen@nt.uni-paderborn.de\vspace{-2em}}

}

\maketitle

\begin{abstract}
Cooperative multi-task semantic communication (CMT-SemCom) improves task execution performance by leveraging shared representations. However, as we demonstrated in \cite{halimi-letter}, cooperative multi-tasking can be either constructive or destructive, depending on the semantic relationships among tasks. To ensure constructive cooperation, we propose a semantic-aware task clustering method for CMT-SemCom. We have formulated a sequential multi-stage optimization problem in which semantically aligned tasks are clustered once after a short initial training phase, and then end-to-end (E2E) joint training is conducted exclusively within the discovered groups. Specifically, the problem decomposes into two stages: (i) a semantic clustering problem leveraging hierarchical density-based spatial clustering, and (ii) an intra-cluster E2E CMT-SemCom learning problem. Simulation results demonstrate that the proposed framework effectively mitigates destructive cooperation and negative transfer, yielding accuracy gains compared to unclustered multi-tasking and individual training baselines.
\end{abstract}


\section{Introduction} \label{section.Intro}

Task-oriented semantic communication (SemCom) is recognized as a key enabler for emerging intelligent wireless networks \cite{Gunduz2022}. In these networks, artificial intelligence (AI)-driven communication extends beyond traditional data-oriented services (e.g., voice and text transmission) to support the execution of intelligent tasks \cite{SemCom6G}.
%
%execution across applications ranging from autonomous driving to smart cities.
%
%Future networks anticipate the development of the Internet of Everything and intelligent applications such as digital twins, which will lead to a dramatic surge in wireless traffic and increased demand for intelligent transmission \cite{SemCom6G-2}. 
%Therefore, task-oriented SemCom focuses on delivering semantic information to achieve various downstream tasks. 
Practical task-oriented communication systems often need to support multiple downstream tasks simultaneously. Therefore, multi-task learning (MTL) strategies have recently been investigated for communication systems to exploit common features across tasks and avoid redundant processing \cite{shao2026}.
%Therefore, recent works have started considering developing multi-task learning (MTL) strategies in the communication system design, exploiting the common features across the tasks \cite{shao2026}.
%To achieve this, early studies consider single-task processing, where the transmitter and receiver are designed to guarantee the extraction and reception of the semantic information and the execution of the task on the receiver side \cite{Shao2021, adaptiveSemCompress,ImportanceAwareSem}. 

MTL has been studied as a mechanism for improving generalization performance through shared representation learning~\cite{caruana1997multitask}. 
%By jointly learning multiple related tasks, MTL enables models to leverage common latent features. 
Consequently, recent works have incorporated MTL into SemCom exclusively based on machine learning (ML) approaches and treating the semantic link as a black-box neural network \cite{10520522,gong2023scalable}. On the other hand, our previous work \cite{halimi-letter} introduces an information maximization (InfoMax) perspective on cooperative multi-task SemCom (CMT-SemCom), moving beyond the black-box use of deep neural networks (DNNs). This work investigates dividing the semantic encoder into a common unit (CU) and multiple specific units (SUs) to enable multi-tasking. The CU captures common information from the observation among multiple tasks, while SUs further process this information for individual tasks. The resulting CMT-SemCom framework facilitates multi-tasking based on a single observation. Subsequently, \cite{halimi-icc, halimi-ojcoms} extended CMT-SemCom to scenarios with distributed partial observations and to rate-limited wireless channels, respectively.
%\cite{10013075,10520522,gong2023scalable}

Despite these advancements, existing multi-task SemCom works, including CMT-SemCom, assume that all tasks are fully related and implicitly learn a shared feature subspace for them. However, applying these methods to unrelated tasks leads to suboptimal task execution and potential performance degradation. As demonstrated in \cite{halimi-letter}, such mismatched cooperation results in \emph{destructive cooperation}, where unrelated tasks negatively impact each other’s performance. This motivates identifying groups of semantically related tasks that should share latent semantic processing via the CU.

%This reveals that efficient multi-task SemCom depends on identifying groups of semantically related tasks that should share latent semantic processing via the CU. 

In MTL, task grouping has traditionally been addressed either through extensive search \cite{MTL:search}, subjective human judgment \cite{MTL:human}, or by exploiting task similarities based on datasets, single-task model parameters, or model gradients \cite{MTL:taskgrouping2,MTL:taskaffinity,MTL:WithWhom,MTL:taskgroup}. These approaches can face significant challenges when adapted to the CMT-SemCom. Specifically, gradient-based or parameter-alignment methods require complex backward passes through the differentiable wireless channel model and multiple encoder-decoder pairs, introducing prohibitive computational overhead during the clustering phase.

Recently, we proposed a lightweight semantic-aware clustering approach based on the empirical probability mass functions of the semantic variables for distributed federated learning (FL)-based multi-user scenarios \cite{halimi-FL-Letter}. The proposed method leveraged an information-theoretic metric, i.e., JS-divergence, to group semantically related tasks while maintaining low communication overhead during the FL process. Such an approach is particularly suitable for distributed SemCom systems, in which encoders are not connected with each other through a CU on the transmitter side, preventing transmission of high-dimensional feature representations, gradients, or intermediate model statistics to a central processor, and guaranteeing communication efficiency.

However, although \cite{halimi-FL-Letter} provides a communication-efficient proxy for task-relatedness, it fails to capture the nuanced, higher-order semantic relationships embedded in the shared representation space. In the centralized CMT-SemCom system, where multi-task training is performed directly on the encoder side with a shared CU, the intermediate CU outputs features that are critical for precise task grouping. Consequently, clustering solely based on the statistics of semantic variables may fail to fully characterize task-relatedness in CMT-SemCom.

To address this issue, we propose an approach to manage the shared subspace, ensuring that only fully or partially related tasks cooperate to guarantee \emph{constructive cooperation} in CMT-SemCom. Specifically, we formulate a multi-stage optimization problem where semantic-aware clustering is performed to identify related tasks, followed by an end-to-end (E2E) learning step that ensures only informative tasks are jointly processed. In particular, we introduce a hybrid clustering pipeline that leverages uniform manifold approximation and projection (UMAP) \cite{mcinnes2018umap} and hierarchical density-based spatial clustering of applications with noise (HDBSCAN) \cite{Campello2013HDBSCAN} to extract task-relatedness from the CU output feature space. Our method operates on the high-dimensional semantic embeddings generated by the shared encoder, capturing complex non-linear relationships between tasks. The clustering is executed once after a short initial training phase, fixing the task groups for the remainder of the training.

In summary, building upon our prior work on the CMT-SemCom, we contribute to a framework capable of mitigating destructive cooperation. Key contributions include:

\begin{itemize}  
    \item Proposing a multi-stage problem for Clustered-CMT-SemCom, integrating semantic-aware clustering to identify related tasks, and an E2E learning step to ensure only informative tasks are jointly processed,
    %\item proposing a general task clustering approach regardless of MTL paradigms for DNNs applicable to both a multi-label domain (same covariate but different labels) and a multi-covariate domain (each task has its own covariate), 
    \item Proposing a task clustering method by applying UMAP and HDBSCAN on extracted semantic features, moving beyond common clustering approaches that rely on data distribution or similarity of models parameters,
    \item Demonstrating that the proposed framework effectively mitigates destructive cooperation and improves task execution performance compared to unclustered multi-tasking and individual task training.
\end{itemize}

\section{Problem Formulation} \label{section.SystemModel}
In this section, we establish a probabilistic model to characterize the semantic source and communication aspects. Based on this foundation, we define the overall Clustered-CMT-SemCom optimization problem.% To make it tractable, we decompose it into two sub-problems: a semantic clustering and a learning sub-problem. If there exists!

\subsection{System Model}\label{subsec.system_prob_model}

We consider a multi-task communication system, consisting of a single transmitter (Tx), multiple receivers (Rxs), and a wireless channel in between. The Tx makes an observation $\bm S$ and extracts multiple task-specific information via a split structure, each corresponding to one of $N$ downstream tasks. At Rx $i$, this information is used to decode a reconstruction of \emph{semantic variable} $Z_i$ to support execution of its task. We denote the tuple $(\bm Z, \bm S)$ as \emph{semantic source} \cite{halimi-letter}, fully  described by its probability distribution $p(\bm{z}, \bm{s})$, where $\bm{Z}=[\,Z_1\, Z_2\,\dots\,Z_N]$.

Our goal is to design an efficient multi-task semantic encoder using data-driven training. In particular, we consider the split encoder structure in Fig.~\ref{fig:clustered_cmt_model}, consisting of $K$ CUs indexed by $k$ and $N$ SUs indexed by $i$. Let $(\bm z, \bm s)$ be a semantic source sample, where only $\bm s$ is available to the Tx. Then, CU $k$ preprocesses this observation $\bm s$ for a subset of semantically related tasks $\mathcal T_k \subseteq \mathcal T = \{ 1, \ldots, N \}$. SU $i\in\mathcal T_k$ takes CU $k$'s output $\bm c_k$ and computes a length-$d$ codeword $\bm x_i$. This codeword is transmitted over an AWGN channel to Rx $i$. The received signal $\bm y_i = \bm x_i + \bm n_i$ with $\bm n_i \sim N(\bm 0_d, \gamma_n^2 \bm I_d)$ is used by decoder $i$ to infer $\hat z_i$, with the goal that $\hat z_i = z_i$. Thus, we have the Markov chain for the $i$-th semantic variable as:\vspace{0.3em}
%\vspace*{-0.2cm}
\begin{equation}
    \begin{aligned}
     p(\bm z_i,\bm{s},\bm{c}_k,\bm{x}_i,\bm{y}_i,\hat{\bm z}_i) =\\[0.0em] & \hspace{-9.8em} p(\bm{s}, \bm z_i)p^{\text{\tiny CU}}_k(\bm{c}_k|\bm{s})p^{\text{\tiny SU}}_i(\bm{x}_i|\bm{c}_k)p(\bm{y}_i|\bm{x}_i)p(\hat{\bm z}_i|\bm{y}_i).
    \end{aligned}
\label{eq:system_probability_markov}
\end{equation}

\begin{figure}
    \centering
    \resizebox{0.5\textwidth}{!}{\definecolor{CU2_color}{RGB}{204,102,119} %{51,34,136}
\definecolor{CU1_color}{RGB}{136,204,238}

\usetikzlibrary{patterns, backgrounds}
\begin{tikzpicture}
        \node[draw, circle, inner sep=0.5pt] (S) {$\begin{tabular}{@{}c@{}}
                                                    \text{Semantic} \\
                                                    \text{Source} \\
                                                    \text{$(\bm{Z}, \bm{S})$}
                                                  \end{tabular}$};

        \node[
    draw=none,
    rectangle,
    right=0.5 cm of S,
    minimum width=1.5cm,
    minimum height=1.8cm,
    inner sep=0pt
] (C) {
    \begin{tabular}{c}
        \rule{0pt}{0.2em}\textbf{CU\textsubscript{1} Enc.} \\
        $\large \vdots$ \\
        \rule{0pt}{1.4em}\textbf{CU\textsubscript{$K^\star$} Enc.}
    \end{tabular}
};

% Hatched rectangles
\begin{scope}
    \begin{pgfonlayer}{background}

        % Top box
        \path[
            draw=black,
            pattern=north west lines,
            pattern color=CU1_color
        ]
        ([xshift=-1.0765cm, yshift=0.4cm]C.center)
        rectangle
        ([xshift=1.0765cm, yshift=0.9cm]C.center);

        % Bottom box
        \path[
            draw=black,
            pattern=north east lines,
            pattern color=CU2_color
        ]
        ([xshift=-1.0765cm, yshift=-0.9cm]C.center)
        rectangle
        ([xshift=1.0765cm, yshift=-0.3cm]C.center);

    \end{pgfonlayer}
\end{scope}

% Dashed side segments (top gap)
\draw[dashed]
    ([xshift=-1.0765cm]C.north)
    --
    ([xshift=-1.0765cm, yshift=0.9cm]C.center);

\draw[dashed]
    ([xshift=1.0765cm]C.north)
    --
    ([xshift=1.0765cm, yshift=0.9cm]C.center);

% Dashed side segments (middle gap)
\draw[dashed]
    ([xshift=-1.0765cm, yshift=0.4cm]C.center)
    --
    ([xshift=-1.0765cm, yshift=-0.3cm]C.center);

\draw[dashed]
    ([xshift=1.0765cm, yshift=0.4cm]C.center)
    --
    ([xshift=1.0765cm, yshift=-0.3cm]C.center);

% Dashed side segments (bottom gap)
\draw[dashed]
    ([xshift=-1.0765cm, yshift=-0.9cm]C.center)
    --
    ([xshift=-1.0765cm]C.south);

\draw[dashed]
    ([xshift=1.0765cm, yshift=-0.9cm]C.center)
    --
    ([xshift=1.0765cm]C.south);
        \node[draw, rectangle, right=0.3 cm of C, yshift=1.1 cm, inner sep=4pt] (SU1) {$\text{SU Enc.}$};
        \node[draw, rectangle, right=0.3 cm of C, yshift=0.0 cm, inner sep=4pt] (SU2) {$\text{SU Enc.}$};
        % Vertical dots between second and third SU Enc.
        \node at ([xshift=1.05cm, yshift=-0.45cm]C.east) {$ \large \vdots$};
        \node[draw, rectangle, right=0.3 cm of C, yshift=-1.1 cm, inner sep=4pt] (SU3) {$\text{SU Enc.}$};
        
        \begin{scope}
            \begin{pgfonlayer}{background}
            % SU1 - North West hatch
            \path[pattern=north west lines, pattern color=CU1_color] 
            ([xshift=-0.75cm, yshift=0.27cm]SU1.center) 
            rectangle 
            ([xshift=0.75cm, yshift=-0.26cm]SU1.center); 

            % SU2 - North East hatch
            \path[pattern=north west lines, pattern color=CU1_color] 
            ([xshift=-0.75cm, yshift=0.27cm]SU2.center) 
            rectangle 
            ([xshift=0.75cm, yshift=-0.26cm]SU2.center);
            \end{pgfonlayer}
        \end{scope}

        \begin{scope}
            \begin{pgfonlayer}{background}
            % SU3 - North East hatch
            \path[pattern=north east lines, pattern color=CU2_color] 
            ([xshift=-0.75cm, yshift=0.27cm]SU3.center) 
            rectangle 
            ([xshift=0.75cm, yshift=-0.26cm]SU3.center); 
            \end{pgfonlayer}
        \end{scope}

        \node[right=0.4 cm of SU1](X1){$\bm{x}_1$};
        \node[right=0.4 cm of SU2](X2){$\bm{x}_2$};
        \node[right=0.4 cm of SU3](Xn){$\bm{x}_N$};

        %channel
        \renewcommand{\arraystretch}{0.85}
        \node[right=of SU3, yshift=1.05 cm, xshift=-0.02 cm] (ch) {$\begin{tabular}{@{}c@{}}
              \footnotesize C\\
              \footnotesize h\\
              \footnotesize a\\
              \footnotesize n\\
              \footnotesize n\\
              \footnotesize e\\
              \footnotesize l
        \end{tabular}$};
        \node[draw, fit=(ch), minimum height=2.0 cm, inner sep=-1pt]{};
        
        \node[draw, rectangle, right=2.5 of SU1] (Rx1) {$\text{Dec.}$};
        \node[draw, rectangle, right=2.5 of SU2] (Rx2) {$\text{Dec.}$};
        \node[draw, rectangle, right=2.5 of SU3] (Rx3) {$\text{Dec.}$};
        % Vertical dots between second and third SU Enc.
        \node at ([xshift=4.75cm, yshift=-0.45cm]C.east) {$ \large \vdots$};
        
        \begin{scope}
            \begin{pgfonlayer}{background}
            % Dec1 - North West hatch
            \path[pattern=north west lines, pattern color=CU1_color] 
            ([xshift=-0.44cm, yshift=0.23cm]Rx1.center) 
            rectangle 
            ([xshift=0.44cm, yshift=-0.23cm]Rx1.center); 

            % Dec2 - North West hatch
            \path[pattern=north west lines, pattern color=CU1_color] 
            ([xshift=-0.44cm, yshift=0.23cm]Rx2.center) 
            rectangle 
            ([xshift=0.44cm, yshift=-0.23cm]Rx2.center); 

            % Dec3 - North East hatch
            \path[pattern=north east lines, pattern color=CU2_color] 
            ([xshift=-0.44cm, yshift=0.23cm]Rx3.center) 
            rectangle 
            ([xshift=0.44cm, yshift=-0.23cm]Rx3.center);
            \end{pgfonlayer}
        \end{scope}
        
        \node[left=0.4 cm of Rx1](X1_hat){$\bm{y}_1$};
        \node[left=0.4 cm of Rx2](X2_hat){$\bm{y}_2$};
        \node[left=0.4 cm of Rx3](Xn_hat){$\bm{y}_N$};
        \node[right=0.4 cm of Rx1](Z1_hat){$\hat{z}_1$};
        \node[right=0.4 cm of Rx2](Z2_hat){$\hat{z}_2$};
        \node[right=0.4 cm of Rx3](Zn_hat){$\hat{z}_N$};

        %\node[above=0.03 cm of C, xshift= 0.7 cm]{$\mathbf{c}$};

        \draw[->, line width=1pt] (S) -- node[above]{\small$\bm{s}$} (C.west);
        \draw[->, line width=1pt] ([yshift=0.6cm]C.east) -- (SU1.west);
        \draw[->, line width=1pt] ([yshift=0.6cm]C.east) -- (SU2.west);
        \draw[->, line width=1pt] ([yshift=-0.6cm]C.east) -- (SU3.west);
        \draw[->, line width=1pt] (SU1) -- (X1);
        \draw[->, line width=1pt] (SU2) -- (X2);
        \draw[->, line width=1pt] (SU3) -- (Xn);
        \draw[->, line width=1pt] (X1_hat) -- (Rx1);
        \draw[->, line width=1pt] (X2_hat) -- (Rx2);
        \draw[->, line width=1pt] (Xn_hat) -- (Rx3);
        \draw[->, line width=1pt] (Rx1) -- (Z1_hat);
        \draw[->, line width=1pt] (Rx2) -- (Z2_hat);
        \draw[->, line width=1pt] (Rx3) -- (Zn_hat);

    \end{tikzpicture}}
    \vspace*{-0.6cm}
    \caption{Illustration of the proposed Clustered-CMT-SemCom.\vspace{-1.5em}}
    \label{fig:clustered_cmt_model}
\end{figure}

\subsection{Problem Statement} \label{subsec.Loss_function}
Previous work \cite{halimi-letter} shows that cooperative multi-tasking is constructive only when semantically related tasks are grouped together. This motivates the split-CU structure we are considering. Following the InfoMax principle, we maximize the mutual information between channel output $\bm Y_i$ and corresponding semantic variable $Z_i$ to design encoder and decoders. That is, for fixed $K$, the joint clustering and InfoMax problem is:
\begin{maxi!} % Maximization environment
    {\substack{
        p^{\text{\tiny CU}}_{1}, \cdots, p^{\text{\tiny CU}}_{K},
        p^{\text{\tiny SU}}_{1}, \cdots, p^{\text{\tiny SU}}_{N},
        \mathcal{T}_1, \cdots, \mathcal{T}_K
    }} % Optimization variables
    {\sum_{k=1}^{K} \sum_{i\in \mathcal{T}_k } I(\bm{Y}_i ; Z_i)} % Objective function
    {\label{eq:main_opt_problem}}{} % Label and optional tag
    \addConstraint{\bigcup_{k=1}^{K} \mathcal{T}_k}{= \mathcal{T}\label{eq:main_opt_problem:c1}} % Constraint 1
	\addConstraint{\mathcal{T}_k \cap \mathcal{T}_{k^\prime}}{= \emptyset, \quad \forall k\neq k^\prime,\label{eq:main_opt_problem:c2}} % Constraint 2
\end{maxi!}
where $I(\bm Y_i;Z_i)$ and $I(\bm Y_j;Z_j)$ for $i,j\in \mathcal{T}_k$, are coupled through the $k$-th CU, $p^{\text{\tiny CU}}_k(\bm c_k|\bm s)$, according to the Markov chain in \eqref{eq:system_probability_markov}. For simplicity in notation, we show $p^{\text{\tiny CU}}_{k}(\bm{c}_k|\bm{s})$ by $p^{\text{\tiny CU}}_{k}$ and also $p^{\text{\tiny SU}}_{i}(\bm{x}_i|\bm{c}_k)$ by $p^{\text{\tiny SU}}_{i}$, representing the distributions for the sub-CUs and SUs, respectively. We observe that constraints \eqref{eq:main_opt_problem:c1} and \eqref{eq:main_opt_problem:c2} ensure that the task clustering is a disjoint partition of $\mathcal T$.

This combines a variational optimization with a clustering problem. A further complication is that the optimal number of clusters $K$ is not known a priori. Jointly optimizing the clustering structure and the variational distributions in \eqref{eq:main_opt_problem} leads to a challenging mixed discrete-continuous optimization. Therefore, in the following section, we design a heuristic procedure to obtain a feasible solution to \eqref{eq:main_opt_problem} with a high objective value.

%This is a variational optimization problem combined with a clustering task. A further complication is that the optimal number $K$ of clusters is not known a priori, and jointly optimizing $K$ and \eqref{eq:main_opt_problem} is intractable. In the following section, we design a heuristic procedure to obtain a feasible solution to \eqref{eq:main_opt_problem} with high objective value.
% To elaborate, a small number of CUs is desirable to harness multi-task learning benefits, while having too few clusters leads to suboptimal performance.

\section{Constructive and Cooperative Multi-Tasking}\label{subsec.system_prob_model}
We approach \eqref{eq:main_opt_problem} and the optimal choice of $K$ through a sequential multi-stage solution approach. First, we jointly determine $K$ and $\mathcal T_1, \ldots, \mathcal T_K$ with a semantics-aware clustering procedure after initial training epochs $E_{\text{init}}$ based on the CMT-SemCom training procedure from \cite{halimi-letter}. Then, we continue training, replacing the initial unified CU with $K^\star$ CUs, exclusively within the task clusters. 

%Both steps are partly data-driven and use a training data set $\mathcal D = \{ \bm s^{(m)}, \bm z^{(m)} \}_{m = 1}^M$ of size $M$.

\subsection{Semantic Clustering}\label{subsec:clustering}
In the first stage, we identify semantically related tasks by directly analyzing the latent feature space of an initially trained unified CU. This allows us to fix the encoder topology before the main cooperative training begins.

\subsubsection{Initial Training} To capture task similarities, we first train a single, unified CU parameterized by $\boldsymbol{\theta}_{\text{u}}$ jointly with all $N$ SUs for a short period of $E_{\text{init}}$ epochs as in~\cite{halimi-letter}. During this bootstrapping phase, the encoder learns to map the source observation $\bm{s}$ to a high-dimensional latent representation $\bm{c} = f_{\boldsymbol{\theta}_{\text{u}}}(\bm{s}) \in \mathbb{R}^{d_{\text{CU}}}$.
To investigate task-relatedness through the CU outputs, we randomly select a representative subset \mbox{$\mathcal D^B_i = \{ (\bm s^{(b)}, z^{(b)}_i)\}_{b = 1}^{B}$} from the complete dataset \mbox{$\mathcal D_i = \{ (\bm s^{(m)}, z^{(m)}_i)\}_{m = 1}^{M_i}$} of each task $i \in \mathcal{T}$. We then define $\mathcal{C}_i$ as the corresponding set of CU outputs for the $i$-th task:

%To capture CU outputs per task to investigate task-relatedness, we use a subset of $B$ random samples from all $N$ tasks, where the dataset for task $i$  is defined as \mbox{$\mathcal D_i = \{ (\bm s^{(b)}, z^{(b)}_i)\}_{b = 1}^{B}$}. We fix $B$ to be large enough to represent the whole dataset and define the set $ \mathcal{C}_i$ to contain the corresponding samples of the CU output for the $i$-th dataset $i \in \mathcal{T}$:

\begin{equation}
    \mathcal{C}_i = \left\{ \bm{c}_i^{(b)}\,|\,  \bm{c}_i^{(b)} = f_{\boldsymbol{\theta}_{\text{u}}}(\bm{s}_i^{(b)}) \, , ( \bm{s}_i^{(b)}, {z}_i^{(b)} ) \in \mathcal D_i \right\}_{b=1}^B.
    \label{eq:cu_feature_set}
\end{equation}

The collection $\mathcal{C} = \bigcup_{i=1}^N \mathcal{C}_i$ contains raw CU outputs that inherently encode task-relevant semantic information shaped by the InfoMax objective in \cite{halimi-letter}.

\subsubsection{Clustering Procedure} We apply Hierarchical Density-Based Spatial Clustering of Applications with Noise (HDBSCAN) \cite{Campello2013HDBSCAN} to cluster tasks based on the extracted semantic features $\mathcal{C}$. We note that, unlike common clustering algorithms such as K-Means, which require specifying the number and shape of the clusters (i.e., a Gaussian assumption in K-Means), HDBSCAN identifies the number of clusters $K^\star$ inherently and assumes no shape for them.

HDBSCAN builds a fully connected graph from all data points, whose edge weights are given by the mutual reachability distance (MRD). The MRD between two points is defined based on their Euclidean distance while accounting for the local density, such that it becomes larger than the Euclidean distance in low-density regions. Next, a minimum spanning tree is constructed from the graph, which is basically hierarchical clustering. Finally, HDBSCAN selects the clusters by maximizing a cluster stability measure, favoring dense clusters that are well separated from each other~\cite{Campello2013HDBSCAN}.
%such that the clusters with the highest density and largest distance from each other are selected~\cite{Campello2013HDBSCAN}. 

%which maximizes the sum over all possible cluster combinations where each cluster is weighted by the inverse difference between smallest and largest MRD, as well as the density of the cluster.

However, directly applying HDBSCAN to high-dimensional CU representations in $\mathbb{R}^{d_{\text{CU}}}$ is ineffective, as it requires more observed samples to produce enough density. To make the density more evident and make clustering easier for HDBSCAN, Uniform Manifold Approximation and Projection (UMAP) \cite{mcinnes2018umap} has been used together with HDBSCAN as a dimensionality reduction technique~\cite{ali2019timecluster}. UMAP preserves the global structure of the data and is designed specifically for clustering in addition to other applications.

% However, directly applying a density-based clustering algorithm, like HDBSCAN, in $\mathbb{R}^{d_{\text{CU}}}$ is susceptible to the curse of dimensionality, which requires more observed samples to produce enough density. To make the density more evident and make clustering easier for HDBSCAN, dimensionality reduction is applied \cite{mcinnes2018umap}.

% Specifically, due to preserving the global structure of the data and being designed for clustering among other duties, Uniform Manifold Approximation and Projection (UMAP) \cite{mcinnes2018umap} has been used together with HDBSCAN to improve clustering performance~\cite{ali2019timecluster}. %pealat2021improved (applied UMAP+HDBSCAN to time series data)   
%\cite{ali2019timecluster,UMAP_HDBSCAN_IEEE}

At a high level, UMAP first constructs a weighted k-neighbour graph to represent the topology of the high-dimensional data, where the weights are the distances between data points scaled by the size of a unit ball to the k-th nearest neighbor. Next, an initial low-dimensional representation estimation is calculated, and an equivalent topological representation is constructed. UMAP then minimizes the cross-entropy between these two topological representations to minimize the error between the low-dimensional and the original high-dimensional one \cite{mcinnes2018umap}.

We apply UMAP directly to the pooled CU outputs $\mathcal{C}$. Let $\Phi_{\text{UMAP}}: \mathbb{R}^{d_{\text{CU}}} \to \mathbb{R}^{d_{\text{low}}}$ denote the learned non-linear projection, where $d_{\text{low}} \ll d_{\text{CU}}$. The projected latent features are given by:
\begin{equation}
    \tilde{\bm{c}}_i^{(b)} = \Phi_{\text{UMAP}}\big(\bm{c}_i^{(b)}\big), \quad \forall i \in \mathcal{T}, \, b=1,\dots,B.
    \label{eq:umap_projection}
\end{equation}

We then apply HDBSCAN to the projected set
$\tilde{\mathcal{C}} = \bigcup_{i=1}^N \tilde{\mathcal{C}}_i$ with
$\tilde{\mathcal{C}}_i = \{\tilde{\bm{c}}_i^{(b)}\}_{b=1}^B$.
%$\tilde{\mathcal{C}} = \{\tilde{\bm{c}}_i^{(b)}\}_{i,b}$ for all $i \in \mathcal T$ and $1 \leq b \leq B$. 
Let $\ell_i^{(b)} \in \{1, \dots, K^\star\}$ denote the cluster label assigned to sample $b$ of task $i$. The final task to cluster mapping $\{\mathcal{T}^\star_1, \dots, \mathcal{T}^\star_{K^\star}\}$ is obtained via majority voting over the sample-level assignments:
\begin{equation}
    \mathcal{T}^\star_k = \left\{ i \in \mathcal{T} \,\middle|\, k=\argmax_{k^\prime=1,\dots,K^\star} \sum_{b=1}^B \mathbb{I}\big(\ell_i^{(b)} = k^\prime\big) \right\},
    \label{eq:task_assignment}
\end{equation}
for $k=1, \dots, K^\star$, and $\mathbb{I}(\ell_i^{(b)} = k^\prime)=1$ if $\ell_i^{(b)} = k^\prime$ and $0$ otherwise. This yields a disjoint partition satisfying $\bigcup_{k=1}^{K^\star} \mathcal{T}^\star_k = \mathcal{T}$ \eqref{eq:main_opt_problem:c1} and $\mathcal{T}^\star_k \cap \mathcal{T}^\star_{k'} = \emptyset$ for $k \neq k'$ \eqref{eq:main_opt_problem:c2}.

Tasks assigned to the same cluster $\mathcal{T}^\star_k$ are deemed semantically compatible for constructive cooperation. In the second stage, the training continues with dedicated sub-CU $k$ assigned to each cluster, replacing the initial unified CU.

\subsection{Clustered-CMT-SemCom Learning}\label{subsec:learning problem}
Having the clusters ($\mathcal{T}^\star_k$) and the number of sub-CUs ($K^\star$) obtained in the first stage, we aim to solve the InfoMax problem by optimizing the SUs and sub-CUs in $K^\star$ parallel learning steps. Therefore, the second-stage optimization problem is defined as:
\vspace{-0.1cm}
\begin{maxi} % Maximization environment
    {\substack{
        p^{\text{\tiny CU}}_{1}, \cdots, p^{\text{\tiny CU}}_{K^\star}, \\[.5em]
        p^{\text{\tiny SU}}_{1}, \cdots, p^{\text{\tiny SU}}_{N} }} % Optimization variables
    {\sum_{k=1}^{K^\star} \sum_{i\in \mathcal{T}^{\star}_k}  I(\bm{Y}_i ; Z_i)} % Objective function
    {\label{eq:second-stage_opt_problem}}{} % Label and optional tag
\end{maxi}

%Expanding the mutual information in our optimization problem, as discussed in detail in Appendix \ref{Appendix:derivation}, the approximated objective function is derived as:
Expanding the mutual information in the optimization problem, as discussed in detail in \cite{halimi-letter}, the approximated objective function is derived as:
%\begin{equation} 
\begin{align}
        &\mathcal{L}(\boldsymbol{\theta}, \boldsymbol{\Phi}, \boldsymbol{\Psi}) = \sum_{k=1}^{K^\star}\sum_{i\in \mathcal{T}^{\star}_k}\, I(\bm{Y}_i; Z_i) \nonumber \\[0.0em]
        &\approx \sum_{k=1}^{K^\star} \underbrace{ \textcolor{black}{\mathbb{E}_{p^{\text{\tiny CU}}_{\boldsymbol{\theta}_k}}\left[\,\sum_{i\in \mathcal{T}^{\star}_k} \overbrace{\textcolor{black}{ \left\{\mathbb{E}_{p(\bm{s},z_i)} \bigg[\,\mathbb{E}_{p^{\text{\tiny SU}}_{\boldsymbol{\phi}_i}} [\,\log q_{\boldsymbol{\psi}_i}(z_i|\bm{y}_i)]\ \bigg]\,\right\} } }^{\text{Task specific processing within each sub-CU}}\right]\ } }_{\text{Intra-cluster cooperation through $k$-th Sub-CU}}. \raisetag{80pt} \label{eq:objective_function}
\end{align}  %  \raisetag{30pt}
%\end{equation}
%\cite{bishop2006pattern}

In \eqref{eq:objective_function}, we employed the variational method using weights in neural networks (NNs) \cite{kingma2013auto}. Thus, the posterior distributions are approximated by NNs, resulting in $p^{\text{\tiny CU}}_{\boldsymbol{\theta}_k}$ and $p^{\text{\tiny SU}}_{\boldsymbol{\phi}_i}$, where $\boldsymbol{\theta}_k$ and $\boldsymbol{\phi}_i$ represent the NN's parameters approximating the $k$-th sub-CU and the $i$-th SU encoders, respectively. Although the $i$-th decoder, $p(z_i|\bm{y}_i)$, can be fully determined using the known distributions, due to the high-dimensional integrals, we approximate it with $q_{\boldsymbol{\psi}_i}(z_i|\bm{y}_i)$, where $\boldsymbol{\psi}_i$ represents NNs approximating the distribution of the $i$-th decoder. Finally, we obtain the empirical estimate of the Clustered-CMT-SemCom objective function for training by approximating the expectations using Monte Carlo sampling, considering the training data set \mbox{$\mathcal D = \{ (\bm s^{(m)}, z^{(m)}_1,\dots,z^{(m)}_N) \}_{m = 1}^{M}$} as done in \cite{halimi-letter}. The overall design of the proposed Clustered-CMT-SemCom framework is summarized and illustrated in Fig. \ref{fig:clustering_sys_model}.
%The technique is widely used in machine learning, e.g., \cite{alemi2016deep}, and also in task-oriented communications, e.g., \cite{Shao2021}.
% \begin{figure}[!t]
%     \centering
%     \resizebox{0.5\textwidth}{!}{\input{Figures/clustering_model.tikz}}
%     \caption{System design illustration of the semantic-aware clustering for constructive-CMT-SemCom.}% The semantic domain is represented by the histogram of semantic variables.}
%     \label{fig:clustering_sys_model}
% \end{figure}
\begin{figure}[!t]
    \centering
    \resizebox{0.5\textwidth}{!}{\input{Figures/clustering_overview.tikz}}
    \vspace*{-0.75cm}
    \caption{Clustered-CMT-SemCom system design illustration.\vspace{-1.5em}}% The semantic domain is represented by the histogram of semantic variables.}
    \label{fig:clustering_sys_model}
\end{figure}

\section{Simulation Results}
To evaluate the performance of Clustered-CMT-SemCom, we compare its task execution error rate against two baselines: Individual training, corresponding to single-task training, and Unclustered-CMT-SemCom, MTL unaware of task-relatedness as in \cite{halimi-letter}. We note that the semantic-aware task clustering method in \cite{halimi-FL-Letter} reduces to Individual training in our considered scenarios since tasks label domains do not overlap. Therefore, it is implicitly included through the Individual training baseline. The simulation code of this work is also available online\footnote{\href{https://github.com/ant-uni-bremen/Semantic_Aware_Task_Clustering}{github.com/ant-uni-bremen/Semantic\_Aware\_Task\_Clustering}}.
%\url{github.com/ant-uni-bremen/Semantic\_Aware\_Task\_Clustering}}.

\subsection{Simulation Setup}\label{Sub:Sim_Setup}
To demonstrate the effectiveness of our approach, we use the dataset $\mathcal{D} = \bigcup_{i=1}^{N} \mathcal{D}_i$ and consider multinomial classification tasks, i.e., each semantic variable has a multinomial distribution \mbox{$Z_i \sim \text{Multinomial}$}. We define the first task T1 on the MNIST dataset of handwritten digits \cite{deng2012mnist} as the classification of $0$ to $9$. For tasks T2 and T3 we consider the EMNIST dataset of handwritten letters \cite{cohen2017emnist} such that T2 (EMNIST-I) is the classification of $A$ to $J$ and T3 (EMNIST-II) is the classification of $K$ to $T$. For T4 and T5 we use the Fashion-MNIST dataset of fashion items \cite{xiao2017fashion}, such that T4 (Fashion-MNIST-I) is the classification of: pullover, dress, coat, sandals, bag, and T5 (Fashion-MNIST-II) is the classification of: t-shirt, trousers, shirt, sneakers, boots. All datasets are grayscale images of size $28\times28$.

%To demonstrate the effectiveness of our approach, we use the dataset $\mathcal{D} = \bigcup_{i=1}^{N} \mathcal{D}_i $ consisting of the MNIST dataset containing handwritten digits \cite{deng2012mnist}, the EMNIST dataset containing handwritten English letters \cite{cohen2017emnist}, and the Fashion-MNIST dataset containing photographs of fashion items \cite{xiao2017fashion}. All datasets are grayscale images of size $28\times28$.

%In our simulations, we consider multinomial classification tasks and therefore, all semantic variables follow a multinomial distribution \mbox{$Z_i \sim \text{Multinomial}$}. We define the first task T1 (MNIST) as the classification of digits $0$ to $9$, the second task T2 (EMNIST-I) as the classification of letters $A$ to $J$, the third task T3 (EMNIST-II) as the classification of letters $K$ to $T$, the fourth task T4 (Fashion-MNIST-I) as the classification of items: pullover, dress, coat, sandals, bag, and the fifth task T5 (Fashion-MNIST-II) for classifying items: t-shirt, trousers, shirt, sneakers, boots. 

To assess the generality of the proposed framework, we evaluate it across different combinations of our defined tasks in three scenarios, with $N=3$ in each, as detailed in Table~\ref{table_scenarios}.

\begin{table}[!t]
\vspace{0.04in}
    \centering
    %\scriptsize
    \renewcommand{\arraystretch}{1.0}
    \caption{List of simulation scenarios.}
    \vspace{-2mm}
    \setlength{\tabcolsep}{1pt}
    \begin{tabular}{p{13mm}|p{16mm} p{27mm}p{29mm} }
        \hline
        \textbf{Scenario}& \multicolumn{3}{l}{\textbf{Tasks}} \\
        \hline
        Scenario A & T1 (MNIST), & T2 (EMNIST-I) , & T4 (Fashion-MNIST-I) \\
        \hline
        Scenario B & T1 (MNIST), & T2 (EMNIST-I), & T3 (EMNIST-II) \\
        \hline
        Scenario C & T1 (MNIST), & T4 (Fashion-MNIST-I), & T5 (Fashion-MNIST-II) \\
        \hline
    \end{tabular}
    \label{table_scenarios}
    \vspace*{-0.0em}
\end{table}

For the semantic clustering stage, we set the initial training period $E_\text{init}=10$ and $B=15000$ samples, i.e., $5000$ samples per task. The UMAP hyperparameters are specified as: the output dimension $d_{\text{low}}=5$, \texttt{min\_dist=0.1}, and \texttt{n\_neighbors=30}. The HDBSCAN  hyperparameters are set as: \texttt{allow\_single\_cluster=True}, \texttt{min\_samples=50}, \texttt{min\_cluster\_size=0.15$\times $5000 = 750 }, \texttt{alpha=1}, and \texttt{cluster\_selection\_epsilon=2}. We use Euclidean distance as the distance metric for both UMAP and HDBSCAN. Furthermore, we merged clusters where a task is dominant in those. By being dominant, we mean that the number of samples of a task in one cluster is greater than $\frac{1}{N}$. This means that if the number of samples from T1 is dominant in multiple clusters, those clusters are merged. 
As a general example, if T1 and T2 are placed into two different clusters containing samples of the same tasks relatively equal and minimal samples of other tasks, then we merge these clusters, concluding that T1 and T2 are semantically related.

The NN architectures in our simulations include Conv2d-based Convolutional Neural Network (CNN) layers with $3\times3$ convolution filters, a stride of $1$, and zero-padding for the encoders, and fully connected layers for the decoders.
Furthermore, 2D MaxPooling of size $2\times2$ for dimensionality reduction, BatchNorm2d, ReLU activations, max-pooling, and dropout layers are used, as summarized in Table~\ref{table_nn_architectures}. We set the batch size to $100$, and use a decaying learning rate strategy with an initial value of $10^{-3}$.

%The presented NN structures represent the general case for a single SU and a single decoder. For each additional task, a corresponding SU and decoder are introduced, ensuring that each task is processed independently within the cooperative framework. For instance, in a three-task scenario, the architecture will include three SUs and three decoders.

\begin{table}[!t]
\setlength{\tabcolsep}{1pt}
    \centering
    \caption{NN architectures used for the compared methods}
    \vspace{-2mm}
    \label{table_nn_architectures}
        \centering
        \begin{tabular}{p{0.05\columnwidth}|p{0.91\columnwidth}}
            \hline
            & \textbf{Layer} \\
            \hline
            \multirow{3}{*}{ \hspace{-1.2mm}\textbf{CU}} & Conv2D (\# filters: $F_1$), BatchNorm, ReLU, MaxPool2D, Dropout ($0.1$) \\ 
             &Conv2D (\# filters: $F_2$), BatchNorm, ReLU, MaxPool2D, Dropout ($0.1$)\\
             &Conv2D (\# filters: $F_3$), BatchNorm, ReLU, MaxPool2D \\
             \hline 
             \multirow{4}{*}{\textbf{SU}} & Conv2D (\# filters: $F_4$), BatchNorm, ReLU \\ 
             &Conv2D (\# filters: $F_5$), BatchNorm, ReLU, Flatten \\
             & Dense (output size: 20), ReLU, Dense (output size: 16), ReLU \\
             & Dense (output size: $d$), ReLU, Power normalization \\
             \hline 
             \multirow{2}{*}{\hspace{-0.2mm}\textbf{Dec}}  & Dense (output size: 16), ReLU, Dense (output size: 16), ReLU  \\ 
             & Dense (output size: number of classes for classification)  \\
             \hline
        \end{tabular}
        \vspace*{-1.0em}
\end{table}

The CU and SU structures provided in Table~\ref{table_nn_architectures} represent the overall system architecture with variables $\bm F= [F_1,\, F_2,\, F_3,\, F_4,\, F_5 ] $, representing the number of convolution filters of the CNN. 
The adjustment of variable $\bm F$ is done so that the total number of trainable parameters remains approximately equal across the Clustered-CMT-SemCom, Unclustered-CMT-SemCom, and Individual training cases. %We note here that the Unclustered-CMT-SemCom is identical in terms of NN architecture to the Clustered-CMT-SemCom if all tasks are clustered together.
%As we have $N=3$ tasks, there exist three possible task groupings. 
In case all tasks are clustered together, which is also the Unclustered-CMT-SemCom, we use $\bm F = [9, 9, 10, 6, 3]$, i.e., the CU for all three tasks followed by three SUs with filters $F_4=6$ and $F_5=3$, resulting in $8745$ trainable parameters.
When two tasks are clustered together and the third one is alone, we use $\bm F = [8, 8, 8, 6, 3]$ for the sub-CU with both clustered tasks and $\bm F = [6, 6, 6, 6, 3]$ for the separate task, integrating the other sub-CU capacity into its SU. This results in a total of $8661$ trainable parameters for all tasks.
%Finally, for the Individual training, we use $\bm F = [6 \,6 \, 6\, 6\, 3]\ $, where the unified CU capacity is integrated into the three separate SUs. In this case the total number of trainable parameters for all tasks is also $8\,661$. Also, we consider an AWGN channel with $d=8$ channel uses for all simulations.
Finally, for Individual training, we use $\bm F = [6, 6, 6, 6, 3]$, integrating the unified CU capacity into the three separate SUs, resulting in $8661$ trainable parameters for all tasks. All simulations consider an AWGN channel with training SNR uniformly distributed between $0$ to $10\,\text{dB}$ and $d=8$ channel uses.

\subsection{Evaluations}
In this section, we first present the performance of the proposed task clustering procedure, then the comparison of Clustered-CMT-SemCom with the aforementioned benchmarks with respect to task execution error rate.  

\subsubsection{Semantic-aware vs. Dataset-based Clustering}
For each scenario, we are comparing the proposed clustering's performance on two cases. It is executed on either (i) the output of the unified CU $\bm c$, as proposed, named \emph{semantic-aware clustering}, or (ii) the raw input image samples $\bm s$, named \emph{Dataset-based clustering}.

As shown in Figs. \ref{fig:cluster_A_umap} to \ref{fig:cluster_C}, with $d_\text{low}=2$ for illustration purpose, clustering performed on raw input data results in Unclustered-CMT-SemCom, where all tasks are jointly trained, for all scenarios. On the other hand, the semantic-aware clustering groups tasks differently in Scenarios A and C. We also observe that in Scenario B, both approaches yield the same clustering result. The impact of these clusters on task execution performance is evaluated in the next section.

\renewcommand{\marksize}{4.2pt}
\newcommand{\marksizeC}{4.4pt}

% ==== COLORS ====
\definecolor{brown1363485}{RGB}{136,34,85}
\definecolor{burlywood221204119}{RGB}{221,204,119}
\definecolor{darkcyan0119187}{RGB}{0,119,187}
\definecolor{darkgray176}{RGB}{176,176,176}
\definecolor{darkcyan0119187}{RGB}{51,34,136}
\definecolor{firebrick2045117}{RGB}{204,51,17}
\definecolor{forestgreen1711951}{RGB}{17,119,51}
\definecolor{indianred204102119}{RGB}{204,102,119}
\definecolor{lightgray204}{RGB}{204,204,204}
\definecolor{olivedrab15315351}{RGB}{153,153,51}
\definecolor{skyblue136204238}{RGB}{136,204,238}

% ==== GLOBAL STYLE KNOBS ====
\pgfplotsset{
    myLineWidth/.style={very thick},
    myMarkerSize/.style={mark size=3},
    myMarkRepeat/.style={mark repeat=50}
}
\newcommand{\subplothight}{4.8cm}

% ==== BASE STYLE ====
\pgfplotsset{
    myBasePlot/.style={
        myLineWidth,
        myMarkerSize,
        myMarkRepeat,
        mark options={solid},
        opacity=1.0
    }
}

\renewcommand{\marksize}{4.2pt}
\newcommand{\linewidthMarker}{0.7pt} %2.04
\newcommand{\sigmaOpacity}{0.0}

% ==== PER-PLOT STYLES (MATCH LEGENDS) ====

% --- Task1 ---
\pgfplotsset{
    SingletaskTask1/.style={
        myBasePlot,
        color=firebrick2045117,
        mark=asterisk,
        solid
    },
    MultitaskTask1/.style={
        myBasePlot,
        color=forestgreen1711951,
        mark=triangle*,
        mark options={rotate=180},
        solid
    },
    ClusteringTask1/.style={
        myBasePlot,
        color=darkcyan0119187,
        mark=square,
        mark options={rotate=45},
        mark size=2.5pt,
        solid
    },
    T1T2t3Task1/.style={
        myBasePlot,
        color=skyblue136204238,
        mark=x,
        solid
    },
    t1T2T3Task1/.style={
        myBasePlot,
        color=brown1363485,
        mark=Mercedes star flipped,  
        solid
    },
    T1t2T3Task1/.style={
        myBasePlot,
        color=olivedrab15315351,
        mark=oplus,
        solid
    },
    DatasetClusterTask1/.style={
        myBasePlot,
        color=olivedrab15315351,
        mark=x,
    },
    DatasetClusterTask1NoMark/.style={
        myBasePlot,
        color=olivedrab15315351,
    },
}

% --- Task2 ---
\pgfplotsset{
    SingletaskTask2/.style={
        myBasePlot,
        color=firebrick2045117,
        mark=asterisk,
        solid
    },
    MultitaskTask2/.style={
        myBasePlot,
        color=forestgreen1711951,
        mark=triangle*,
        mark options={rotate=180},
        solid
    },
    ClusteringTask2/.style={
        myBasePlot,
        color=darkcyan0119187,
        mark=square,
        mark options={rotate=45},
        mark size=2.5pt,
        solid
    },
    T1T2t3Task2/.style={
        myBasePlot,
        color=skyblue136204238,
        mark=x,
        solid
    },
    t1T2T3Task2/.style={
        myBasePlot,
        color=brown1363485,
        mark=Mercedes star flipped,  
        solid
    },
    T1t2T3Task2/.style={
        myBasePlot,
        color=olivedrab15315351,
        mark=oplus,
        solid
    },
    DatasetClusterTask2/.style={
        myBasePlot,
        color=olivedrab15315351,
        mark=x,
    },
    DatasetClusterTask2NoMark/.style={
        myBasePlot,
        color=olivedrab15315351,
    },
}

% --- Task3 ---
\pgfplotsset{
    SingletaskTask3/.style={
        myBasePlot,
        color=firebrick2045117,
        mark=asterisk,
        solid
    },
    MultitaskTask3/.style={
        myBasePlot,
        color=forestgreen1711951,
        mark=triangle*,
        mark options={rotate=180},
        solid
    },
    ClusteringTask3/.style={
        myBasePlot,
        color=darkcyan0119187,
        mark=square,
        mark options={rotate=45},
        mark size=2.5pt,
        solid
    },
    T1T2t3Task3/.style={
        myBasePlot,
        color=skyblue136204238,
        mark=x,
        solid
    },
    t1T2T3Task3/.style={
        myBasePlot,
        color=brown1363485,
        mark=Mercedes star flipped,  
        solid
    },
    T1t2T3Task3/.style={
        myBasePlot,
        color=olivedrab15315351,
        mark=oplus,
        solid
    },
    DatasetClusterTask3/.style={
        myBasePlot,
        color=olivedrab15315351,
        mark=x,
    },
    DatasetClusterTask3NoMark/.style={
        myBasePlot,
        color=olivedrab15315351,
    },
}

% \begin{figure}[!t]
%     \centering
%     \scalebox{0.9}{\input{Figures/clustering_results/over_epochs_scenario_A_subplots.tikz}}
%     \caption{
%     Task execution performance comparison: Scenario A
%     }
%     \label{fig:overEpochsA}
% \end{figure}

% \begin{figure}[!t]
%     \centering
%     \scalebox{0.9}{\input{Figures/clustering_results/over_epochs_scenario_B_subplots.tikz}}
%     \caption{
%     Task execution performance comparison: Scenario B
%     }
%     \label{fig:overEpochsB}
% \end{figure}

% \begin{figure}[!t]
%     \centering
%     \scalebox{0.9}{\input{Figures/clustering_results/over_epochs_scenario_C_subplots.tikz}}
%     \caption{
%     Task execution performance comparison: Scenario C
%     }
%     \label{fig:overEpochsC}
% \end{figure}

\subsubsection{Task Execution Performance}
We compare the performance of Clustered-CMT-SemCom, representing the semantic-aware clustering, against Unclustered-CMT-SemCom, which coincides here with the dataset-based clustering result, where all tasks are jointly trained, and Individual training, which is simply the single-task training. All results show the average performance of ten independent Monte Carlo runs at $5\,\text{dB}$ SNR.
%which can also represent

As shown in Figs. \ref{fig:overEpochsA} and \ref{fig:overEpochsC}, the proposed Clustered-CMT-SemCom achieves a significantly lower task execution error rate for all tasks compared to Unclustered-CMT-SemCom. Furthermore, the negative transfer is obvious in these scenarios, where Unclustered-CMT-SemCom performs worse than Individual training.  This means MTL for all tasks together is not efficient. In addition, the Clustered-CMT-SemCom outperforms the Individual training on average, either performing better than or matching single-task training. We note that the match is because the corresponding task is clustered alone.

For Scenario B, we consider an additional case where we group the tasks intuitively named as \emph{Intuitive clustering}, where T2 and T3 are clustered together since they come from the same dataset, and T1 is trained individually. Fig. \ref{fig:overEpochsB} shows that the proposed framework outperforms the Individual training and also the Intuitive clustering on average. As the semantic-aware clustering groups all tasks together, we observe that Unclustered-CMT-SemCom has the same performance as Clustered-CMT-SemCom.

Finally, we evaluate the performance of the Clustered-CMT-SemCom over an SNR range from $-5$ to $20\,\text{dB}$ as shown in Fig.~\ref{fig:overSNR_A} for Scenario A. The advantage of the clustering approach is more apparent at higher SNRs, as for lower SNRs the performance is limited by the limited channel capacity for all approaches, and this holds also for Scenarios B and C.

%Finally, for the task T4 (Fashion-MNIST-I), the performance of the Clustered-CMT-SemCom matches the performance of the Unclustered-CMT-SemCom until $E_\text{init}=10$ epochs (indicated by a black dotted line), and thereafter it approaches the performance of the Individual training, which makes sense, as T4 is a separate cluster from T1 and T2 in the Clustered-CMT-SemCom.

%This shows that if all tasks are similar having one CU is most advantageous, with our approach of Clustered-CMT-SemCom automatically determining if all tasks should be clustered together.

%Finally, we show the task execution error rate over the channel SNR for scenario A in Fig. \ref{fig:overSNR_A}. For lower SNR all approaches perform nearly the same, as the bad channel conditions dominate the task execution performance. For higher SNR, the proposed Clustered-CMT-SemCom approach outperforms the other methods.

\begin{figure}[!t]
    \centering
    \vspace{-2.5mm}
    \begin{subfigure}{0.49\linewidth}
        \centering
        \caption{Semantic-aware clustering}
        \includegraphics[width=\linewidth]
        {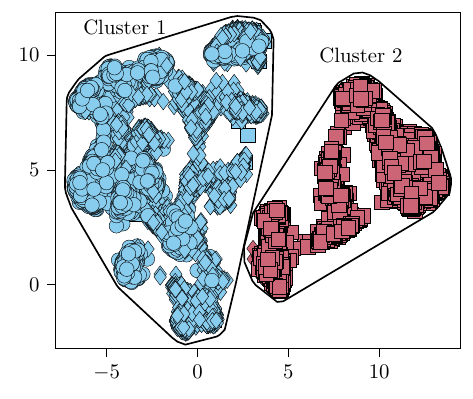}
        \label{fig:cluster_A_umap_1}
    \end{subfigure}
    \hspace{-5mm}
    \begin{subfigure}{0.49\linewidth}
        \centering
        \caption{Dataset-based clustering}
        \includegraphics[width=\linewidth]
        {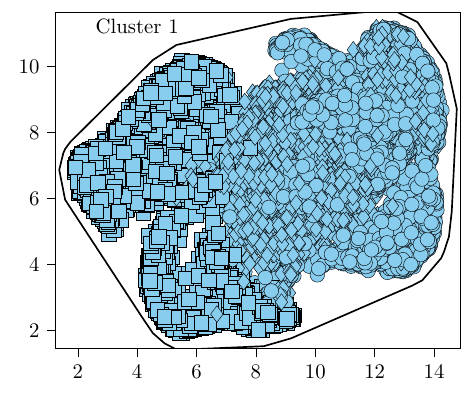}
        \label{fig:cluster_A_umap_2}
    \end{subfigure}
    %\par\vspace{-2mm}
    % \begin{subfigure}{0.49\linewidth}
    %     \centering
    %     \caption{Clustering results CU output + t-SNE}
    %     \includegraphics[width=\linewidth]
    %     {Figures/clustering_results/Scenario_A_clustering_TSNE.pdf}
    % \end{subfigure}
    % \hspace{-5mm}
    % \begin{subfigure}{0.49\linewidth}
    %     \centering
    %     \caption{Clustering results raw data + t-SNE}
    %     \includegraphics[width=\linewidth]
    %     {Figures/clustering_results/Scenario_A_clustering_data_TSNE.pdf}
    % \end{subfigure}
    \par\vspace{-5mm}
    \begin{subfigure}{\linewidth}
        \centering
        \begin{tikzpicture}

% ==== COLORS ====

	\definecolor{brown1363485}{RGB}{136,34,85}
	\definecolor{burlywood221204119}{RGB}{221,204,119}
	\definecolor{darkcyan0119187}{RGB}{0,119,187}
	\definecolor{darkgray176}{RGB}{176,176,176}
	\definecolor{darkslateblue5134136}{RGB}{51,34,136}
	\definecolor{firebrick2045117}{RGB}{204,51,17}
	\definecolor{forestgreen1711951}{RGB}{17,119,51}
	\definecolor{indianred204102119}{RGB}{204,102,119}
	\definecolor{lightgray204}{RGB}{204,204,204}
	\definecolor{olivedrab15315351}{RGB}{153,153,51}
	\definecolor{skyblue136204238}{RGB}{136,204,238}

\begin{axis}[
    hide axis,
    xmin=0, xmax=1,
    ymin=0, ymax=1,
    scale only axis,
    axis lines=none,
    width=4cm,
    height=1.2cm,
    legend style={
        draw=lightgray204,
        fill=white,
        inner sep=2pt,
        outer sep=0pt,
        font=\small,
        legend columns=3,
        column sep=1.5pt,
        /tikz/every even column/.append style={column sep=9.7pt}
    },
    legend cell align=left,
]

% \addlegendimage{burlywood221204119, mark = o, mark size=\marksize,line width=\linewidthMarker,only marks}
% \addlegendentry{MNIST}

% \addlegendimage{darkcyan0119187, mark =o, mark size=\marksize,line width=\linewidthMarker,only marks}
% \addlegendentry{EMNIST I}

% \addlegendimage{brown1363485, mark = o, mark size=\marksizeC,line width=\linewidthMarker,only marks}
% \addlegendentry{Fashion-MNIST I}

\addlegendimage{draw=black,fill=white, mark = *, mark size=\marksize,line width=\linewidthMarker,only marks}
\addlegendentry{T1 (MNIST)}

\addlegendimage{draw=black,fill=white, mark = diamond*, mark size=\marksize,line width=\linewidthMarker,only marks}
\addlegendentry{T2 (EMNIST-I)}

\addlegendimage{draw=black,fill=white, mark = square*, mark size=\marksize,line width=\linewidthMarker,only marks}
\addlegendentry{T4 (Fashion-MNIST-I)}

\end{axis}
\end{tikzpicture}
    \end{subfigure}
    \par\vspace{-7mm}
    \vspace{-5mm}
    \caption{Semantic-aware vs. dataset-based: Scenario A.
    }
    \label{fig:cluster_A_umap}
\end{figure}

\begin{figure}[!t]
\vspace{-1mm}
    \centering
    \vspace{-2.5mm}
    \begin{subfigure}{0.49\linewidth}
        \centering
        \caption{Semantic-aware clustering}
        \includegraphics[width=\linewidth]
        {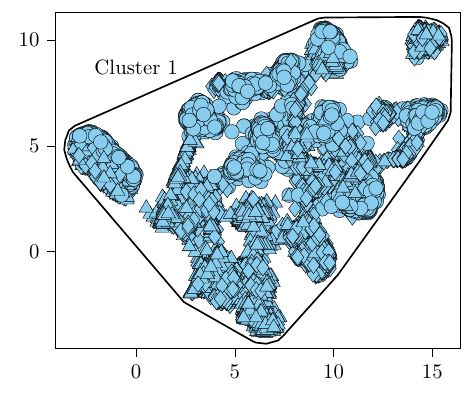}
        \label{fig:cluster_B_1}
    \end{subfigure}
    \hspace{-5mm}
    \begin{subfigure}{0.49\linewidth}
        \centering
        \caption{Dataset-based clustering}
        \includegraphics[width=\linewidth]
        {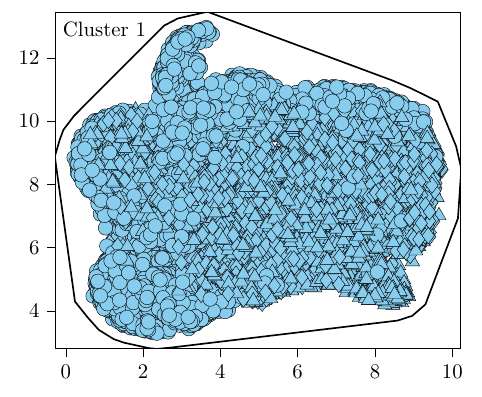}
        \label{fig:cluster_B_2}
    \end{subfigure}
    \par\vspace{-5mm}
    \begin{subfigure}{\linewidth}
        \centering
        \begin{tikzpicture}

% ==== COLORS ====

	\definecolor{brown1363485}{RGB}{136,34,85}
	\definecolor{burlywood221204119}{RGB}{221,204,119}
	\definecolor{darkcyan0119187}{RGB}{0,119,187}
	\definecolor{darkgray176}{RGB}{176,176,176}
	\definecolor{darkslateblue5134136}{RGB}{51,34,136}
	\definecolor{firebrick2045117}{RGB}{204,51,17}
	\definecolor{forestgreen1711951}{RGB}{17,119,51}
	\definecolor{indianred204102119}{RGB}{204,102,119}
	\definecolor{lightgray204}{RGB}{204,204,204}
	\definecolor{olivedrab15315351}{RGB}{153,153,51}
	\definecolor{skyblue136204238}{RGB}{136,204,238}

\begin{axis}[
    hide axis,
    xmin=0, xmax=1,
    ymin=0, ymax=1,
    scale only axis,
    axis lines=none,
    width=4cm,
    height=1.2cm,
    legend style={
        draw=lightgray204,
        fill=white,
        inner sep=2pt,
        outer sep=0pt,
        font=\small,
        legend columns=3,
        column sep=2pt,
        /tikz/every even column/.append style={column sep=10pt}
    },
    legend cell align=left,
]

% \addlegendimage{darkcyan0119187, mark = o, mark size=\marksize,line width=\linewidthMarker,only marks}
% \addlegendentry{EMNIST I}

% \addlegendimage{skyblue136204238, mark = o, mark size=\marksize,line width=\linewidthMarker,only marks}
% \addlegendentry{EMNIST II}

% \addlegendimage{burlywood221204119, mark = o, mark size=\marksize,line width=\linewidthMarker,only marks}
% \addlegendentry{MNIST}

\addlegendimage{draw=black,fill=white, mark = *, mark size=\marksize,line width=\linewidthMarker,only marks}
\addlegendentry{T1 (MNIST)}

\addlegendimage{draw=black,fill=white, mark = diamond*, mark size=\marksize,line width=\linewidthMarker,only marks}
\addlegendentry{T2 (EMNIST-I)}

\addlegendimage{draw=black,fill=white, mark = triangle*, mark size=\marksize,line width=\linewidthMarker,only marks}
\addlegendentry{T3 (EMNIST-II)}

\end{axis}
\end{tikzpicture}
    \end{subfigure}
    \par\vspace{-7mm}
    \caption{Semantic-aware vs. dataset-based: Scenario B.
    }
    \label{fig:cluster_B}
\end{figure}

\begin{figure}[!t]
\vspace{-1mm}
    \centering
    \vspace{-2.5mm}
    \begin{subfigure}{0.49\linewidth}
        \centering
        \caption{Semantic-aware clustering}
        \includegraphics[width=\linewidth]
        {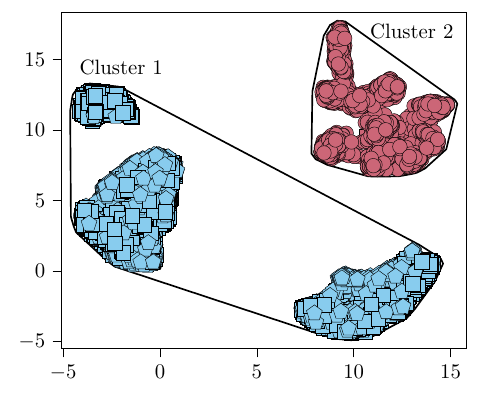}
        \label{fig:cluster_C_1}
    \end{subfigure}
    \hspace{-5mm}
    \begin{subfigure}{0.49\linewidth}
        \centering
        \caption{Dataset-based clustering}
        \includegraphics[width=\linewidth]
        {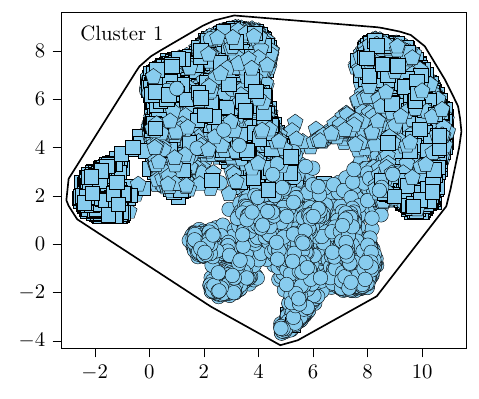}
        \label{fig:cluster_C_2}
    \end{subfigure}
    \par\vspace{-5mm}
    \begin{subfigure}{\linewidth}
        \centering
        \begin{tikzpicture}

% ==== COLORS ====

	\definecolor{brown1363485}{RGB}{136,34,85}
	\definecolor{burlywood221204119}{RGB}{221,204,119}
	\definecolor{darkcyan0119187}{RGB}{0,119,187}
	\definecolor{darkgray176}{RGB}{176,176,176}
	\definecolor{darkslateblue5134136}{RGB}{51,34,136}
	\definecolor{firebrick2045117}{RGB}{204,51,17}
	\definecolor{forestgreen1711951}{RGB}{17,119,51}
	\definecolor{indianred204102119}{RGB}{204,102,119}
	\definecolor{lightgray204}{RGB}{204,204,204}
	\definecolor{olivedrab15315351}{RGB}{153,153,51}
	\definecolor{skyblue136204238}{RGB}{136,204,238}

\begin{axis}[
    hide axis,
    xmin=0, xmax=1,
    ymin=0, ymax=1,
    scale only axis,
    axis lines=none,
    width=4cm,
    height=1.2cm,
    legend style={
        draw=lightgray204,
        fill=white,
        inner sep=2pt,
        outer sep=0pt,
        font=\small,
        legend columns=3,
        column sep=0.5pt,
        /tikz/every even column/.append style={column sep=5pt}
    },
    legend cell align=left,
]

% \addlegendimage{brown1363485, mark = o, mark size=\marksize,line width=\linewidthMarker,only marks}
% \addlegendentry{Fashion-MNIST I}

% \addlegendimage{indianred204102119, mark = o, mark size=\marksize,line width=\linewidthMarker,only marks}
% \addlegendentry{Fashion-MNIST II}

% \addlegendimage{burlywood221204119, mark = o, mark size=\marksize,line width=\linewidthMarker,only marks}
% \addlegendentry{MNIST}

\addlegendimage{draw=black,fill=white, mark = *, mark size=\marksize,line width=\linewidthMarker,only marks}
\addlegendentry{T1 ({\footnotesize MNIST})}

\addlegendimage{draw=black,fill=white, mark = square*, mark size=\marksize,line width=\linewidthMarker,only marks}
\addlegendentry{T4 ({\footnotesize Fashion-MNIST-I})}

\addlegendimage{draw=black,fill=white, mark = pentagon*, mark size=\marksize,line width=\linewidthMarker,only marks}
\addlegendentry{T5 ({\footnotesize Fashion-MNIST-II})}

\end{axis}
\end{tikzpicture}
    \end{subfigure}
    \par\vspace{-7mm}
    \vspace{-5mm}
    \caption{Semantic-aware vs. dataset-based: Scenario C.\vspace{-1.0em}}
    \label{fig:cluster_C}
\end{figure}

%\FloatBarrier

\begin{figure*}[!t]
    \centering
    \hspace{-0.2cm}
    \begin{subfigure}[t]{0.33\textwidth}
        \centering
        \captionsetup{
        margin={0.6cm,0cm}
        } \caption{Scenario A}
        \vspace*{-0.17cm}
            \input{Figures/clustering_results/over_epochs_scenario_A_subplots.tikz}
        \label{fig:overEpochsA}
    \end{subfigure}%
    \hspace{0.16cm}
    \begin{subfigure}[t]{0.33\textwidth}
        \centering
        \caption{Scenario B}
            \input{Figures/clustering_results/over_epochs_scenario_B_subplots.tikz}
        \label{fig:overEpochsB}
    \end{subfigure}%
    \begin{subfigure}[t]{0.33\textwidth}
        \centering
        \caption{Scenario C}
        \vspace*{-0.1cm}
            \input{Figures/clustering_results/over_epochs_scenario_C_subplots.tikz}
        \label{fig:overEpochsC}
    \end{subfigure}
    \captionsetup{
    skip=-12pt}
    \caption{Task execution error rate over the training epochs for the three compared scenarios.\vspace{-1.5em}}
    \label{fig:overEpochs}
\end{figure*}

% \begin{figure*}[!t]
%     \centering
%     \hspace{-0.2cm}
%     \begin{subfigure}[t]{0.33\textwidth}
%         \centering
%         \captionsetup{
%         margin={0.6cm,0cm}
%         } \caption{Scenario A}
%         \vspace*{-0.2cm}
%             \input{Figures/clustering_results/over_epochs_scenario_A_subplots_REVISION.tikz}
%         \label{fig:overEpochsA}
%     \end{subfigure}%
%     \hspace{0.16cm}
%     \begin{subfigure}[t]{0.33\textwidth}
%         \centering
%         \caption{Scenario B}
%             \input{Figures/clustering_results/over_epochs_scenario_B_subplots_REVISION.tikz}
%         \label{fig:overEpochsB}
%     \end{subfigure}%
%     \begin{subfigure}[t]{0.33\textwidth}
%         \centering
%         \caption{Scenario C}
%         \vspace*{-0.1cm}
%             \input{Figures/clustering_results/over_epochs_scenario_C_subplots_REVISION.tikz}
%         \label{fig:overEpochsC}
%     \end{subfigure}
%     \captionsetup{
%     skip=-12pt}
%     \caption{Task execution error rate over the training epochs for the three compared scenarios.\vspace{-1.5em}}
%     \label{fig:overEpochs}
% \end{figure*}

% \begin{figure}[t]
%     \centering
%     \scalebox{0.9}{\input{Figures/clustering_results/over_SNR_scenario_B_subplots.tikz}}
%     \caption{
%     Scenario B
%     }
%     \label{fig:ssim-psnr}
% \end{figure}

% \begin{figure}[t]
%     \centering
%     \scalebox{0.9}{\input{Figures/clustering_results/over_SNR_scenario_C_subplots.tikz}}
%     \caption{
%     Scenario C
%     }
%     \label{fig:ssim-psnr}
% \end{figure}

%-------- OVER SNR Results-----------%
\pgfplotsset{
    myLineWidth/.style={very thick},
    myMarkerSize/.style={mark size=3},
    myMarkRepeat/.style={mark repeat=3},
}
\renewcommand{\marksize}{4.2pt}

% \begin{figure}[!t]
%    \centering
%    \scalebox{0.9}{\input{Figures/clustering_results/over_SNR_scenario_A_subplots.tikz}}
%    \caption{
%    Task performance over SNR for Scenario A.
%    }
%    \label{fig:overSNR_A}
% \end{figure}

\renewcommand{\subplothight}{4.4cm}
\begin{figure}[!t]
   \centering
   \scalebox{0.9}{\begin{tikzpicture}

\begin{groupplot}[
group style={
    group size=1 by 3,
    horizontal sep=1.5cm,
    vertical sep = 2pt,
},
width=0.5\textwidth,
height=\subplothight,
log basis y={10},
ymode=log,
tick align=outside,
tick pos=left,
x grid style={darkgray176},
y grid style={darkgray176},
xmin=-5, xmax=20,
ymin=0.0133124534780709, ymax=0.576813327927558,
xlabel={Epoch},
ylabel={Error rate},
xmajorgrids,
ymajorgrids,
yminorgrids,
minor y tick num=1,
y grid style={darkgray176},
ytick={0.01,0.02,0.03,0.04,0.05,0.06,0.07,0.08,0.09,0.1,0.2,0.3,0.4,0.5,0.6,0.7,0.9,1},
xtick={-5,0,5,10,15,20},
xticklabels=\empty,
    xlabel={},
    xtick align=inside,
 yticklabels={
  \(\displaystyle {0.01}\),\(\displaystyle {0.02}\),,,\(\displaystyle {0.05}\),,,,,
  \(\displaystyle {0.1}\),\(\displaystyle {0.2}\),,,\(\displaystyle {0.5}\),,,,,
  \(\displaystyle {10^{0}}\),,,,,,,,,
  \(\displaystyle {10^{1}}\)
},
ylabel style={yshift=0.2cm},
legend style={draw=lightgray204,at={(0.999,0.999)},
    anchor=north east}
]

% ================= TASK 1 =================
\nextgroupplot[ymin=0.011,ymax=0.46]

\node[
    fill=white,
    draw,
    align=center
] at (axis cs:-1.0,0.021) {T1 (MNIST)};

\addplot [SingletaskTask1]
table {
-10 0.687099993228912
-9 0.65336000919342
-8 0.613180041313171
-7 0.569620013237
-6 0.514200031757355
-5 0.456220000982285
-4 0.390399992465973
-3 0.321360051631927
-2 0.254060000181198
-1 0.188659995794296
0 0.130699992179871
1 0.0870399922132492
2 0.0530200004577637
3 0.0328600108623505
4 0.0230000074952841
5 0.0189000014215708
6 0.0168000105768442
7 0.0159199945628643
8 0.0153799951076508
9 0.0151199940592051
10 0.0148799894377589
11 0.0147000076249242
12 0.0143399955704808
13 0.014219993725419
14 0.0142000019550323
15 0.0141799990087748
16 0.0141799990087748
17 0.014139992184937
18 0.0140999909490347
19 0.0141199948266149
20 0.0141000030562282
};
\addlegendentry{Individual training}

\addplot [MultitaskTask1]
table {
-10 0.686340034008026
-9 0.650560021400452
-8 0.610979974269867
-7 0.566240012645721
-6 0.511979937553406
-5 0.452099978923798
-4 0.388300031423569
-3 0.322800010442734
-2 0.256060034036636
-1 0.189139991998672
0 0.133080005645752
1 0.0885799899697304
2 0.0561400055885315
3 0.0363199934363365
4 0.0266200006008148
5 0.0227200090885162
6 0.0207600053399801
7 0.0198599938303232
8 0.0193399973213673
9 0.0187799986451864
10 0.018559992313385
11 0.0184599999338388
12 0.0182600021362305
13 0.0180400125682354
14 0.0179999954998493
15 0.0177600029855967
16 0.017680000513792
17 0.0177000053226948
18 0.0176999978721142
19 0.0175599996000528
20 0.0176200028508902
};
\addlegendentry{Unclustered-CMT-SemCom}

\addplot [ClusteringTask1]
table {
-10 0.684920006990433
-9 0.649360007047653
-8 0.608279991149902
-7 0.563560009002686
-6 0.510260003805161
-5 0.449860006570816
-4 0.386540001630783
-3 0.319320005178452
-2 0.252159994840622
-1 0.185020005702972
0 0.127120000123978
1 0.0816599905490875
2 0.0496799945831299
3 0.0297799944877625
4 0.019979989528656
5 0.0165199935436249
6 0.0148599982261658
7 0.0140600025653839
8 0.0137799978256226
9 0.0133400022983551
10 0.013019996881485
11 0.012720000743866
12 0.0125200033187866
13 0.012419992685318
14 0.0123600006103516
15 0.0124000012874603
16 0.0122800052165985
17 0.012199991941452
18 0.0120799958705902
19 0.0121199905872345
20 0.0120999991893768
};
\addlegendentry{Clustered-CMT-SemCom}

% ================= TASK 2 =================
\nextgroupplot[ymin=0.039,ymax=0.48,xtick={-5,0,5,10,15}]

\node[
    fill=white,
    draw,
    align=center
] at (axis cs:8,0.22) {T2 (EMNIST-I)};

\addplot [SingletaskTask2]
table {
-10 0.693539977073669
-9 0.660119950771332
-8 0.621820032596588
-7 0.577880024909973
-6 0.528219997882843
-5 0.471939980983734
-4 0.409740030765533
-3 0.34568002820015
-2 0.28150001168251
-1 0.221079990267754
0 0.165319994091988
1 0.120420001447201
2 0.0888800024986267
3 0.0689399987459183
4 0.0576799996197224
5 0.0514000058174133
6 0.0485999993979931
7 0.0465999953448772
8 0.0451600067317486
9 0.0443800017237663
10 0.0438200011849403
11 0.0434600003063679
12 0.0432799980044365
13 0.042899988591671
14 0.0427599921822548
15 0.0426400080323219
16 0.0426400117576122
17 0.0426400005817413
18 0.0424400046467781
19 0.0425399951636791
20 0.0425200052559376
};

\addplot [MultitaskTask2]
table {
-10 0.697440028190613
-9 0.663760006427765
-8 0.624680042266846
-7 0.582100033760071
-6 0.532539963722229
-5 0.475820004940033
-4 0.415380001068115
-3 0.351319998502731
-2 0.287239998579025
-1 0.226240009069443
0 0.171979993581772
1 0.129139974713326
2 0.0980000048875809
3 0.0765800029039383
4 0.064640000462532
5 0.0591399893164635
6 0.0554399974644184
7 0.0539200007915497
8 0.0528399869799614
9 0.0519799999892712
10 0.0514599978923798
11 0.0510400049388409
12 0.0503400042653084
13 0.0500800013542175
14 0.049940001219511
15 0.0497799888253212
16 0.0494600050151348
17 0.0493200048804283
18 0.0491200014948845
19 0.0490000136196613
20 0.0488600023090839
};

\addplot [ClusteringTask2]
table {
-10 0.693060004711151
-9 0.659340006113052
-8 0.620859998464584
-7 0.576899993419647
-6 0.525119996070862
-5 0.467720013856888
-4 0.406840008497238
-3 0.340839999914169
-2 0.276259994506836
-1 0.214579993486404
0 0.161259996891022
1 0.117839998006821
2 0.085999995470047
3 0.0644199967384338
4 0.0540199995040894
5 0.0482999980449677
6 0.0455199956893921
7 0.0438199996948242
8 0.0429000020027161
9 0.0420799970626831
10 0.0417999982833862
11 0.0415200054645538
12 0.0412200093269348
13 0.0411199986934662
14 0.0409200012683868
15 0.0407400012016296
16 0.0406200051307678
17 0.0405800044536591
18 0.0404399871826172
19 0.0404600024223328
20 0.04028000831604
};

% ================= TASK 3 =================
\nextgroupplot[ymin=0.04,ymax=0.37,xtick={-5,0,5,10,15,20},
xticklabels={-5,0,5,10,15,20},
xtick align=outside,
    xlabel={$\text{SNR}\,[\text{dB}]$},]

\node[
    fill=white,
    draw,
    align=center
] at (axis cs:8,0.26) {T4 (Fashion-MNIST-I)};

\addplot [SingletaskTask3]
table {
-10 0.540359973907471
-9 0.505499958992004
-8 0.465859979391098
-7 0.421240031719208
-6 0.373459994792938
-5 0.326800018548965
-4 0.275539994239807
-3 0.224139973521233
-2 0.176439985632896
-1 0.137000009417534
0 0.10393999516964
1 0.0810199975967407
2 0.0657799988985062
3 0.0582199990749359
4 0.0536000020802021
5 0.0519000068306923
6 0.0510000064969063
7 0.0505999848246574
8 0.049979992210865
9 0.0498999953269958
10 0.0496200025081635
11 0.0493000037968159
12 0.0489199943840504
13 0.0485799983143806
14 0.0484799928963184
15 0.0485999882221222
16 0.0485800020396709
17 0.0484999939799309
18 0.048640001565218
19 0.0485999993979931
20 0.0485799983143806
};

\addplot [MultitaskTask3]
table {
-10 0.563360035419464
-9 0.529679954051971
-8 0.491400003433228
-7 0.452179998159409
-6 0.410519987344742
-5 0.364100009202957
-4 0.318259954452515
-3 0.271120011806488
-2 0.22877998650074
-1 0.191679999232292
0 0.160279989242554
1 0.137280002236366
2 0.12109999358654
3 0.110519990324974
4 0.105379983782768
5 0.101839996874332
6 0.0999199971556664
7 0.0984999984502792
8 0.0976400002837181
9 0.0968599915504456
10 0.0963800027966499
11 0.0957000106573105
12 0.0951000079512596
13 0.0946799963712692
14 0.0943000093102455
15 0.0940600037574768
16 0.0939199924468994
17 0.0937999933958054
18 0.0937200039625168
19 0.0936999917030334
20 0.0936600044369698
};

\addplot [ClusteringTask3]
table {
-10 0.541279995441437
-9 0.505619996786118
-8 0.466160005331039
-7 0.421380001306534
-6 0.373019993305206
-5 0.323220002651215
-4 0.273479998111725
-3 0.224160003662109
-2 0.177880001068115
-1 0.137399995326996
0 0.105620002746582
1 0.0810399949550629
2 0.066159999370575
3 0.0572800099849701
4 0.0529600024223328
5 0.0511799991130829
6 0.0501799941062927
7 0.0495000004768372
8 0.04932000041008
9 0.0491200029850006
10 0.0488399982452393
11 0.0484999895095825
12 0.0482199966907501
13 0.0480199933052063
14 0.0478999972343445
15 0.0480000019073486
16 0.0479400038719177
17 0.0478399932384491
18 0.0478799939155579
19 0.0478200018405914
20 0.047819995880127
};

Fashion\end{groupplot}

\end{tikzpicture}}
   \captionsetup{
    skip=-3pt}
   \caption{
   Scenario A: Results over SNR \vspace{-1.5em}
   }
   \label{fig:overSNR_A}
\end{figure}

% Clustering Plots
\pgfplotsset{
    myLineWidth/.style={very thick},
    myMarkerSize/.style={mark size=3},
    myMarkRepeat/.style={mark repeat=50}
}
\pgfplotsset{
    myBasePlot/.style={
        myLineWidth,
        myMarkerSize,
        myMarkRepeat,
        mark options={solid},
        opacity=1.0
    }
}

\renewcommand{\subplothight}{4.5cm}
\begin{figure*}[!t]
    \centering
    \begin{tikzpicture}
\begin{axis}[
    hide axis, scale only axis, width=1pt, height=1pt, xmin=0, xmax=1, ymin=0, ymax=1,
    legend columns=6,
    legend style={font=\footnotesize, draw=lightgray204, at={(0.5,0.5)}, anchor=center, /tikz/every even column/.append style={column sep=0.27cm}},
]
\addlegendimage{SingletaskTask1}
\addlegendentry{$\!\left\{\text{T1}\right\},\left\{\text{T2}\right\}, \left\{\text{T4}\right\}$}
\addlegendimage{MultitaskTask1}
\addlegendentry{$\!\left\{\text{T1},\!\text{T2},\! \text{T4}\right\}$}
\addlegendimage{T1T2t3Task1}
\addlegendentry{$\!\left\{\text{T1},\!\text{T2}\right\}, \left\{\text{T4}\right\}$ }
\addlegendimage{T1t2T3Task1}
\addlegendentry{$\!\left\{\text{T1},\!\text{T4}\right\},\left\{\text{T2}\right\}$ }
\addlegendimage{t1T2T3Task1}
\addlegendentry{$\!\left\{\text{T2},\!\text{T4}\right\},\left\{\text{T1}\right\}$ }
\addlegendimage{ClusteringTask1}
\addlegendentry{\!Clustered-CMT-SemCom \!\!}
\end{axis}
\end{tikzpicture}\par
    \vspace{-0.5cm}
    \hspace{-0.2cm}
    \begin{subfigure}[t]{0.33\textwidth}
        \centering
        \captionsetup{
        margin={0.6cm,0cm}
        } \caption{T1 (MNIST)}
        \vspace*{-0.1cm}
            \input{Figures/clustering_results/over_epochs_scenario_A_subplots_REVISION_T1.tikz}
        \label{fig:overEpochsAGrid}
    \end{subfigure}%
    \hspace{0.16cm}
    \begin{subfigure}[t]{0.33\textwidth}
        \centering
        \caption{T2 (EMNIST-I)}
        \vspace*{-0.24cm}
            \input{Figures/clustering_results/over_epochs_scenario_A_subplots_REVISION_T2.tikz}
        \label{fig:overEpochsBGrid}
    \end{subfigure}%
    \begin{subfigure}[t]{0.33\textwidth}
        \centering
        \caption{T4 (Fashion-MNIST-I)}
        \vspace*{-0.24cm}
        %\vspace*{-0.1cm}
            \input{Figures/clustering_results/over_epochs_scenario_A_subplots_REVISION_T4.tikz}
        \label{fig:overEpochsCGrid}
    \end{subfigure}
    \captionsetup{
    skip=-16pt}
    \caption{\vspace{-0.2em}Task performance comparison between the exhaustive-search clustering and Clustered-CMT-SemCom for Scenario A.\vspace{-1.5em}}
    \label{fig:overEpochsGridSearch}
\end{figure*}

\begin{figure}[!t]
   \centering
   \scalebox{0.9}{% clustering sweep: share of the runs with the correct clustering (mean over the scenarios) for every value of every parameter, the
% other parameters at their clustering_params value (circles if SWEEP_FIGURE_SELECTED: value of clustering_params). Written by plot_all.py
\begin{tikzpicture}
\providecolor{darkcyan0119187}{RGB}{51,34,136}
\providecolor{firebrick2045117}{RGB}{204,51,17}
\providecolor{darkgray176}{RGB}{176,176,176}
\providecolor{lightgray204}{RGB}{204,204,204}
\pgfplotsset{
    sweepHDBSCANcuoutput/.style={color=darkcyan0119187, mark=*, mark size=1.5pt, thick},  % Semantic-aware clustering
    sweepHDBSCANrawdata/.style={color=firebrick2045117, mark=*, mark size=1.5pt, semithick, mark options={fill=white}},  % Dataset-based clustering
    sweepused/.style={only marks, mark=o, mark size=3pt, black, thin},  % value of clustering_params
    sweepbar/.style={draw=#1, fill=#1, bar width=7pt},
}
\begin{groupplot}[
    group style={group size=2 by 4, horizontal sep=0.15cm, vertical sep=1.45cm, yticklabels at=edge left, ylabels at=edge left},
    scale only axis, width=(\columnwidth-2.05cm)/2, height=1.3cm,
    ymin=25, ymax=100, ytick={0,25,50,75,100}, ylabel={Correct [\%]},
    clip mode=individual, every axis title shift=1pt, title style={font=\small},  % markers at the border not clipped; title close to the axis
    xmajorgrids, ymajorgrids, grid style={darkgray176}, tick align=outside, tick pos=left,
]

\nextgroupplot[xmin=-0.5, xmax=5.5, xtick={0,1,2,3,4,5}, xticklabels={10,20,30,40,50,60},
    title={\texttt{n\_neighbors}}]
\addplot [sweepHDBSCANcuoutput] coordinates {(0,95.33) (1,99.33) (2,99.33) (3,98) (4,98) (5,95.33)};
\addplot [sweepHDBSCANrawdata] coordinates {(0,77.33) (1,59.33) (2,36.67) (3,33.33) (4,33.33) (5,33.33)};

\nextgroupplot[xmin=-0.5, xmax=4.5, xtick={0,1,2,3,4}, xticklabels={0,0.05,0.1,0.15,0.2},
    title={\texttt{min\_dist}}]
\addplot [sweepHDBSCANcuoutput] coordinates {(0,96) (1,98) (2,99.33) (3,97.33) (4,94)};
\addplot [sweepHDBSCANrawdata] coordinates {(0,60) (1,45.33) (2,36.67) (3,33.33) (4,33.33)};

\nextgroupplot[xmin=-0.5, xmax=5.5, xtick={0,1,2,3,4,5}, xticklabels={1,2,3,4,5,6},
    title={$d_{\text{low}}$}]
\addplot [sweepHDBSCANcuoutput] coordinates {(0,34) (1,96.67) (2,99.33) (3,98) (4,99.33) (5,99.33)};
\addplot [sweepHDBSCANrawdata] coordinates {(0,33.33) (1,33.33) (2,34.67) (3,38.67) (4,36.67) (5,36.67)};

\nextgroupplot[xmin=-0.5, xmax=6.5, xtick={0,1,2,3,4,5,6}, xticklabels={1,5,10,25,50,75,100},
    title={\texttt{min\_samples}}, xticklabel style={rotate=90, anchor=east}]
\addplot [sweepHDBSCANcuoutput] coordinates {(0,99.33) (1,99.33) (2,99.33) (3,99.33) (4,99.33) (5,99.33) (6,99.33)};
\addplot [sweepHDBSCANrawdata] coordinates {(0,36.67) (1,36.67) (2,36.67) (3,36.67) (4,36.67) (5,36.67) (6,36.67)};

\nextgroupplot[xmin=-0.5, xmax=5.5, xtick={0,1,2,3,4,5}, xticklabels={0.05,0.1,0.15,0.2,0.25,0.3},
    title={relative\\[-3pt]\texttt{min\_cluster\_size}}, title style={align=center}, xticklabel style={rotate=90, anchor=east}]
\addplot [sweepHDBSCANcuoutput] coordinates {(0,98.67) (1,98.67) (2,99.33) (3,98.67) (4,95.33) (5,74)};
\addplot [sweepHDBSCANrawdata] coordinates {(0,36.67) (1,36.67) (2,36.67) (3,36.67) (4,36.67) (5,36.67)};

\nextgroupplot[xmin=-0.5, xmax=2.5, xtick={0,1,2}, xticklabels={0.1,0.5,1},
    title={\texttt{alpha}}]
\addplot [sweepHDBSCANcuoutput] coordinates {(0,91.33) (1,94.67) (2,99.33)};
\addplot [sweepHDBSCANrawdata] coordinates {(0,64) (1,45.33) (2,36.67)};

\nextgroupplot[xmin=-0.5, xmax=6.5, xtick={0,1,2,3,4,5,6}, xticklabels={0,0.5,1,1.5,2,2.5,3},
    title={\texttt{cluster\_selection\_epsilon}}]
\addplot [sweepHDBSCANcuoutput] coordinates {(0,96.67) (1,96.67) (2,96.67) (3,98) (4,99.33) (5,98) (6,97.33)};
\addplot [sweepHDBSCANrawdata] coordinates {(0,60) (1,60) (2,52.67) (3,42) (4,36.67) (5,34) (6,33.33)};

\end{groupplot}

\begin{axis}[
    at={(group c1r4.south east)}, anchor=south west, xshift=0.15cm, scale only axis, width={(\columnwidth-2.05cm)/2-0.65cm}, height=1.3cm,
    ybar, xmin=-0.5, xmax=0.5, xtick={0}, xticklabels={Best setting},
    ymin=0, ymax=100, ytick={0,25,50,75,100}, ytick pos=right, yticklabel pos=right, xtick pos=left,
    ymajorgrids, grid style={darkgray176}, tick align=outside, xtick style={draw=none},
    legend to name=sweeplegend, legend columns=2, legend style={font=\footnotesize, draw=lightgray204, /tikz/every even column/.append style={column sep=6pt}},
    legend cell align=left,  % legend: below the figure
]
\addlegendimage{sweepHDBSCANcuoutput, sharp plot, line legend}
\addlegendentry{Semantic-aware clustering}
\addlegendimage{sweepHDBSCANrawdata, sharp plot, line legend}
\addlegendentry{Dataset-based clustering}
\addplot [sweepbar=darkcyan0119187, bar shift=-3.5pt, forget plot] coordinates {(0,99.33)};
\addplot [sweepbar=firebrick2045117, bar shift=3.5pt, forget plot] coordinates {(0,77.33)};
\end{axis}

% legend centered below the figure, close to the lowest tick labels
\node[anchor=north, yshift=-1pt] at (current bounding box.south) {\pgfplotslegendfromname{sweeplegend}};

\end{tikzpicture}}
   \captionsetup{
    skip=-4pt}
    \caption{\vspace{-0.0em}Impact of hyperparameters on clustering.\vspace{-1.5em}}
   \label{fig:parameter_investigation}
\end{figure}

\subsection{Discussion}

\subsubsection{Semantic-aware vs. Exhaustive-search Clustering}
In Fig.~\ref{fig:overEpochsGridSearch}, we compare the performance of the semantic-aware clustering with the exhaustive search for Scenario A for a training and evaluation SNR of $20\,\text{dB}$. For $N=3$, there are $5$ possible combinations of clustering. As shown, Clustered-CMT-SemCom performs the same as the optimum of the exhaustive search, which is $\left\{\text{T1},\!\text{T2}\right\}\!,\! \left\{\text{T4}\right\}$, being the same clustering obtained by Clustered-CMT-SemCom.

\subsubsection{Clustering hyperparameter sensitivity study}

We investigate the impact of UMAP and HDBSCAN hyperparameters mentioned in Section \ref{Sub:Sim_Setup} on clustering in Fig.~\ref{fig:parameter_investigation}.
The parameters specified in Section \ref{Sub:Sim_Setup} serve as a baseline, with each parameter varied individually in Fig.~\ref{fig:parameter_investigation}.
We show their impact on clustering accuracy, which is the percentage of cases where the clustering output matches the optimum found by exhaustive-search clustering for $50$ Monte Carlo runs of each Scenario A, B, and C. %meaning combined $150$ runs.

\subsubsection{Runtime analysis}
Finally, we mention that the runtime for performing the proposed clustering, based on the CPU time measurement, resulted in about $0.3\%$ of the total time it takes to cluster and train for $250$ epochs. This emphasizes efficiency of Clustered-CMT-SemCom compared to exhaustive search.

\section{Conclusion} \label{section.Conclusion}
We proposed the Clustered-CMT-SemCom framework that uses semantic-aware clustering to mitigate destructive cooperation between unrelated tasks by formulating a sequential multi-staged optimization problem. The proposed system adaptively assigns semantically related tasks to shared subspaces, ensuring constructive cooperation. We demonstrated that our approach improves task performance in various scenarios over different datasets, where benchmarks like Individual training and Unclustered-CMT-SemCom suffer due to neglect of statistical relationships among semantic variables.

\bibliographystyle{IEEEtran}
\bibliography{IEEEabrv,References.bib}

\begin{thebibliography}{10}
\providecommand{\url}[1]{#1}
\csname url@samestyle\endcsname
\providecommand{\newblock}{\relax}
\providecommand{\bibinfo}[2]{#2}
\providecommand{\BIBentrySTDinterwordspacing}{\spaceskip=0pt\relax}
\providecommand{\BIBentryALTinterwordstretchfactor}{4}
\providecommand{\BIBentryALTinterwordspacing}{\spaceskip=\fontdimen2\font plus
\BIBentryALTinterwordstretchfactor\fontdimen3\font minus \fontdimen4\font\relax}
\providecommand{\BIBforeignlanguage}[2]{{%
\expandafter\ifx\csname l@#1\endcsname\relax
\typeout{** WARNING: IEEEtran.bst: No hyphenation pattern has been}%
\typeout{** loaded for the language `#1'. Using the pattern for}%
\typeout{** the default language instead.}%
\else
\language=\csname l@#1\endcsname
\fi
#2}}
\providecommand{\BIBdecl}{\relax}
\BIBdecl

\bibitem{halimi-letter}
A.~Halimi~Razlighi, C.~Bockelmann, and A.~Dekorsy, ``Semantic communication for cooperative multi-task processing over wireless networks,'' \emph{IEEE Wireless Communications Letters}, vol.~13, no.~10, pp. 2867--2871, 2024.

\bibitem{Gunduz2022}
D.~Gündüz, Z.~Qin, I.~E. Aguerri, H.~S. Dhillon, Z.~Yang, A.~Yener, K.~K. Wong, and C.-B. Chae, ``Beyond transmitting bits: Context, semantics, and task-oriented communications,'' \emph{IEEE Journal on Selected Areas in Communications}, vol.~41, no.~1, pp. 5--41, 2023.

\bibitem{SemCom6G}
C.~You, Y.~Cai, Y.~Liu, M.~Di~Renzo, T.~M. Duman, A.~Yener, and A.~L. Swindlehurst, ``Next generation advanced transceiver technologies for 6{G} and beyond,'' \emph{IEEE Journal on Selected Areas in Communications}, vol.~43, no.~3, pp. 582--627, 2025.

\bibitem{shao2026}
H.~Li, S.~Xie, J.~Shao, Z.~Wang, H.~He, S.~Song, J.~Zhang, and K.~B. Letaief, ``Mutual information-empowered task-oriented communication: Principles, applications and challenges,'' \emph{IEEE Communications Magazine}, vol.~64, no.~4, pp. 164--171, 2026.

\bibitem{caruana1997multitask}
R.~Caruana, ``Multitask learning,'' \emph{Machine learning}, vol.~28, pp. 41--75, 1997.

\bibitem{10520522}
Y.~E. Sagduyu, T.~Erpek, A.~Yener, and S.~Ulukus, ``Multi - receiver task-oriented communications via multi - task deep learning,'' in \emph{2023 IEEE Future Networks World Forum (FNWF)}, 2023, pp. 1--6.

\bibitem{gong2023scalable}
M.~Gong, S.~Wang, and S.~Bi, ``A scalable multi-device semantic communication system for multi-task execution,'' in \emph{GLOBECOM 2023-2023 IEEE Global Communications Conference}.\hskip 1em plus 0.5em minus 0.4em\relax IEEE, 2023, pp. 2227--2232.

\bibitem{halimi-icc}
A.~Halimi~Razlighi, M.~H.~V. Tillmann, E.~Beck, C.~Bockelmann, and A.~Dekorsy, ``Cooperative and collaborative multi-task semantic communication for distributed sources,'' in \emph{ICC 2025 - IEEE International Conference on Communications}, 2025, pp. 3966--3971.

\bibitem{halimi-ojcoms}
A.~Halimi~Razlighi, C.~Bockelmann, and A.~Dekorsy, ``Semantic communication for cooperative multi-tasking over rate-limited wireless channels with implicit optimal prior,'' \emph{IEEE Open Journal of the Communications Society}, vol.~6, pp. 8523--8538, 2025.

\bibitem{MTL:search}
T.~Standley, A.~Zamir, D.~Chen, L.~Guibas, J.~Malik, and S.~Savarese, ``Which tasks should be learned together in multi-task learning?'' in \emph{Proceedings of the 37th International Conference on Machine Learning}, ser. ICML'20.\hskip 1em plus 0.5em minus 0.4em\relax JMLR.org, 2020.

\bibitem{MTL:human}
Y.~Zhang and Q.~Yang, ``A survey on multi-task learning,'' \emph{IEEE Transactions on Knowledge and Data Engineering}, vol.~34, no.~12, pp. 5586--5609, 2022.

\bibitem{MTL:taskgrouping2}
Y.~Wei, Z.~Hu, L.~Shen, Z.~Wang, Y.~Li, C.~Yuan, and D.~Tao, ``Task groupings regularization: data-free meta-learning with heterogeneous pre-trained models,'' in \emph{Proceedings of the 41st International Conference on Machine Learning}, ser. ICML'24.\hskip 1em plus 0.5em minus 0.4em\relax JMLR.org, 2024.

\bibitem{MTL:taskaffinity}
A.~Ayman, A.~Mukhopadhyay, and A.~Laszka, ``Task grouping for automated multi-task machine learning via task affinity prediction,'' \emph{arXiv preprint arXiv:2310.16241}, 2023.

\bibitem{MTL:WithWhom}
Z.~Kang, K.~Grauman, and F.~Sha, ``Learning with whom to share in multi-task feature learning,'' in \emph{Proceedings of the 28th International Conference on International Conference on Machine Learning}, ser. ICML'11.\hskip 1em plus 0.5em minus 0.4em\relax Madison, WI, USA: Omnipress, 2011, p. 521–528.

\bibitem{MTL:taskgroup}
C.~Fifty, E.~Amid, Z.~Zhao, T.~Yu, R.~Anil, and C.~Finn, ``Efficiently identifying task groupings for multi-task learning,'' in \emph{Proceedings of the 35th International Conference on Neural Information Processing Systems}.\hskip 1em plus 0.5em minus 0.4em\relax Red Hook, NY, USA: Curran Associates Inc., 2021.

\bibitem{halimi-FL-Letter}
\BIBentryALTinterwordspacing
A.~H. Razlighi, P.~Dhingra, E.~Beck, B.~Matthiesen, and A.~Dekorsy, ``Semantic-aware task clustering for federated cooperative multi-task semantic communication,'' 2026. [Online]. Available: \url{https://arxiv.org/abs/2601.17419}
\BIBentrySTDinterwordspacing

\bibitem{mcinnes2018umap}
L.~McInnes, J.~Healy, and J.~Melville, ``{UMAP}: Uniform manifold approximation and projection for dimension reduction,'' \emph{arXiv preprint arXiv:1802.03426}, 2018.

\bibitem{Campello2013HDBSCAN}
R.~J. G.~B. Campello, D.~Moulavi, and J.~Sander, ``Density-based clustering based on hierarchical density estimates,'' in \emph{Advances in Knowledge Discovery and Data Mining}, J.~Pei, V.~S. Tseng, L.~Cao, H.~Motoda, and G.~Xu, Eds.\hskip 1em plus 0.5em minus 0.4em\relax Berlin, Heidelberg: Springer Berlin Heidelberg, 2013, pp. 160--172.

\bibitem{ali2019timecluster}
M.~Ali, M.~W. Jones, X.~Xie, and M.~Williams, ``Timecluster: dimension reduction applied to temporal data for visual analytics,'' \emph{The Visual Computer}, vol.~35, no.~6, pp. 1013--1026, 2019.

\bibitem{kingma2013auto}
D.~P. Kingma and M.~Welling, ``Auto-encoding variational bayes,'' \emph{arXiv preprint arXiv:1312.6114}, 2013.

\bibitem{deng2012mnist}
L.~Deng, ``The mnist database of handwritten digit images for machine learning research,'' \emph{IEEE Signal Processing Magazine}, vol.~29, no.~6, pp. 141--142, 2012.

\bibitem{cohen2017emnist}
G.~Cohen, S.~Afshar, J.~Tapson, and A.~Van~Schaik, ``Emnist: Extending mnist to handwritten letters,'' in \emph{2017 international joint conference on neural networks (IJCNN)}.\hskip 1em plus 0.5em minus 0.4em\relax IEEE, 2017, pp. 2921--2926.

\bibitem{xiao2017fashion}
H.~Xiao, K.~Rasul, and R.~Vollgraf, ``Fashion-mnist: a novel image dataset for benchmarking machine learning algorithms,'' \emph{arXiv preprint arXiv:1708.07747}, 2017.

\end{thebibliography}

\end{document}